\documentclass{article} 
\usepackage{iclr2027_conference,times}

\newif\ifarxiv
\arxivtrue

\usepackage{amsmath,amsfonts,bm}

\def\eqref#1{equation~\ref{#1}}

\def\1{\bm{1}}

\DeclareMathAlphabet{\mathsfit}{\encodingdefault}{\sfdefault}{m}{sl}
\SetMathAlphabet{\mathsfit}{bold}{\encodingdefault}{\sfdefault}{bx}{n}

\usepackage[table]{xcolor}
\usepackage{tabularx}
\definecolor{oursbg}{HTML}{D6E6F5}
\usepackage{hyperref}
\usepackage{url}
\usepackage{colortbl}
\usepackage{graphicx}
\usepackage{wrapfig}
\usepackage{booktabs}
\usepackage{algorithm}
\usepackage{multirow}
\usepackage{wrapfig}
\usepackage{algpseudocode}
\usepackage{caption}
\graphicspath{{figures/}}

\title{SteerQuant: Steering Quantization Error with Action-Guided Scaling in World–Action Models}

\ifarxiv
\iclrfinalcopy
\newcommand{\affiliationOne}{Tongji}
\newcommand{\affiliationTwo}{HKUST}
\newcommand{\affiliationThree}{HIT}
\newcommand{\affiliationFour}{EPFL}

\author{%
\begin{minipage}[t]{\dimexpr\textwidth-2\tabcolsep\relax}
\centering\normalfont\small
\textbf{Yunhan Wang}$^{1,2,*,\dagger}$ \hspace{0.75em}%
\textbf{Haodong Wang}$^{2,*}$ \hspace{0.75em}%
\textbf{Zhiming Liu}$^{3}$ \hspace{0.75em}%
\textbf{Zicong Hong}$^{4}$ \hspace{0.75em}%
\textbf{Qianli Liu}$^{2}$ \\[0.2em]
\textbf{Xiaoyi Pang}$^{2}$ \hspace{0.75em}%
\textbf{Yangjia Hu}$^{2}$ \hspace{0.75em}%
\textbf{Quanxin Shou}$^{2}$ \hspace{0.75em}%
\textbf{Yikun Miao}$^{2}$ \hspace{0.75em}%
\textbf{Song Guo}$^{2}$ \\[0.65em]
\footnotesize
$^{1}$\affiliationOne \hspace{1.25em}%
$^{2}$\affiliationTwo \hspace{1.25em}%
$^{3}$\affiliationThree \hspace{1.25em}%
$^{4}$\affiliationFour\\[0.65em]
Project page: \url{https://cathyyw.github.io/Steerquant-Website/}
\end{minipage}%
}

\hypersetup{
    pdftitle={SteerQuant: Steering Quantization Error with Action-Guided Scaling in World--Action Models},
    pdfauthor={Yunhan Wang, Haodong Wang, Zhiming Liu, Zicong Hong, Qianli Liu, Xiaoyi Pang, Yangjia Hu, Quanxin Shou, Yikun Miao, Song Guo}
}

\else
\author{Anonymous authors}
\hypersetup{
  pdftitle={SteerQuant: Steering Quantization Error with Action-Guided Scaling in World--Action Models},
  pdfauthor={}
}
\fi

\newcommand{\methodname}{SteerQuant}

\begin{document}

\maketitle

\ifarxiv
\fancyhead{}
\renewcommand{\headrulewidth}{0pt}

\begingroup
\renewcommand{\thefootnote}{\fnsymbol{footnote}}
\footnotetext[1]{Equal contribution.}
\footnotetext[2]{Work done during Yunhan Wang's internship at HKUST.}
\endgroup

\fi

\begin{abstract}
World--action models (WAMs) jointly generate future world states
and actions through iterative denoising, using shared weights
to process heterogeneous semantic streams of video, proprioceptive,
and action tokens.
Quantization reduces inference cost, but comparable numerical
errors in different streams can have markedly different effects
on final actions, making numerical accuracy alone insufficient
for reliable control.
We introduce \textbf{SteerQuant}, a 4-bit quantization framework
for WAMs that steers errors toward computations with less
influence on final actions.
It maps how each stream's quantization errors affect final
actions and uses this map to guide shared channel scaling.
Activation scaling is further calibrated for each stream and
denoising step to accommodate changes in activation ranges
and action impact.
This adapts quantization to different stream requirements
without duplicating weights or increasing bit-widths
for selected streams.
To reduce the extra kernel launches and memory traffic
introduced by scaling, we develop \textbf{Rudder}, a 4-bit
inference engine for WAMs that fuses scaling and output
compensation into low-bit kernels.
{At W4A8 and W4A4, SteerQuant maintains mean LIBERO success rates within
0.8 percentage points of BF16 on both Cosmos-Policy and FastWAMJoint.
On RoboCasa tasks with Cosmos-Policy, SteerQuant improves mean success
over the strongest evaluated baseline at each precision by 4.59 percentage
points at W4A8 and 5.55 at W4A4.
Across three WAMs, deployment with Rudder reduces peak GPU
memory and achieves up to $2.23\times$ DiT denoising speedup over BF16.}
On a real dual-arm robot, W4A8 deployment achieves a
$1.35\times$ end-to-end inference speedup while maintaining
average task success relative to BF16.

\end{abstract}

\section{Introduction}
World--action models (WAMs) build on pretrained video models
to jointly model future world states and actions, using
spatiotemporal priors learned from large-scale video
data~\citep{kim2026cosmos,agarwal2026cosmos,ye2026world,yuan2026fast}.
During inference, diffusion Transformers~\citep{peebles2023scalable} repeatedly process
heterogeneous semantic streams, including visual representations,
proprioceptive states, and actions, across denoising steps.
This repeated computation and data movement make inference costly.

Post-training quantization (PTQ) reduces these costs~\citep{frantar2022gptq,li2023q,shang2023post,hu2026mosaicquant} by lowering
weight and activation precision, reducing memory traffic and
enabling hardware-accelerated low-precision arithmetic (e.g., INT4 and NVFP4).
To preserve accuracy, existing PTQ methods typically 
control outlier magnitudes to reduce numerical 
error~\citep{xiao2023smoothquant,ashkboos2024quarot,li2025svdquant,liu2025spinquant}. 
However, WAMs use shared weights to process heterogeneous 
semantic streams, so a single weight quantization choice 
affects multiple streams at once. 
The resulting errors propagate through stream interactions 
and subsequent denoising steps, potentially producing 
different final-action deviations even at similar magnitudes. 
Effective WAM quantization must therefore balance errors 
across streams according to their impact on final actions.




Achieving this balance requires identifying which streams'
errors most affect final actions.
We observe that comparable numerical errors across semantic
streams can lead to markedly different final-action deviations
and task success rates (see Sec.~\ref{sec:preliminary}).
A naive method is to calibrate weight--activation scaling
independently for each stream.
However, independently chosen scales generally require
different quantized weight matrices, sacrificing weight sharing.
Moreover, a stream whose errors strongly affect final actions
at one denoising step may have less influence at another,
limiting the effectiveness of fixed per-stream choices.
The core challenge is therefore to adapt error trade-offs
across streams as their action impact changes, while retaining
one static quantized weight matrix per layer and fixed bit-widths.

To address this challenge, we propose \textbf{\methodname{}},
a 4-bit quantization framework for WAMs that steers errors
toward computations with less influence on final actions.
It coordinates shared channel scaling and stream-specific
activation scaling across denoising steps while retaining
shared weights and fixed 4-bit.
To reduce the extra kernel launches and memory traffic
introduced by separate scaling operations, we fuse these
transformations into low-bit kernels.

Our contributions are summarized as follows:
\begin{itemize}
    \item We reveal that heterogeneous semantic streams differ
    in their tolerance to quantization: comparable numerical
    errors can produce markedly different action deviations
    and task success rates. This tolerance varies across layers
    and denoising steps, motivating calibration guided by
    downstream action impact.

    \item We propose \textbf{\methodname{}}, which uses an Action-Impact
    Stream Map to guide shared channel scaling and stream-specific
    activation calibration across denoising steps. These two stages
    prioritize computations with greater action impact while
    retaining one static quantized weight matrix per layer
    and fixed bit-widths.

    \item We develop \textbf{Rudder}, a 4-bit inference engine
    for WAMs that integrates calibrated scaling and output
    compensation into fused kernels. By reducing extra kernel
    launches and intermediate memory traffic, Rudder enables
    efficient execution of \methodname{} while preserving
    the computational and memory benefits of quantization.

    \item {We evaluate three WAMs across LIBERO, RoboLab, and RoboCasa.
SteerQuant preserves near-BF16 success on LIBERO. On RoboCasa tasks
with Cosmos-Policy, it improves success over the strongest evaluated
same-precision baseline by up to 5.55 percentage points.
With Rudder, we achieve up to $2.23\times$ DiT denoising speedup over BF16.
Real dual-arm deployment delivers a $1.35\times$ end-to-end inference
speedup while maintaining average task success.}
\end{itemize}

\section{Related Work}
\textbf{Post-Training Quantization.}
Existing PTQ methods broadly follow two directions.
The first reduces numerical error through transformations~\citep{lin2023awq,shao2024omniquant,sun2025flatquant,wang2026twinquant},
such as SmoothQuant's equivalent channel
scaling~\citep{xiao2023smoothquant}.
The second incorporates action-related criteria into
quantization~\citep{xu2026qvla}.
For VLA models, ActQuant combines action-guided bit allocation
with weight-scale optimization~\citep{akbari2026actquant}.
For WAMs, QuantWAMs allocates weight precision using sensitivity
to the joint video--action loss~\citep{zhou2026quantwams}.
However, these criteria do not directly characterize how
weight and activation quantization errors in coupled WAM
streams propagate to final actions across denoising steps.
We explore this impact to guide calibration at fixed 4-bit. {We additionally compare with a fixed-W4 adaptation of QuantWAMs that retains its activation protections (Appendix~\ref{app:additional_baselines}).}

\textbf{Efficient WAM Inference.}
Joint future-state and action prediction makes iterative
WAM denoising costly.
Existing acceleration methods reduce this cost by simplifying
future prediction or reusing intermediate computations.
For example, Fast-WAM retains video prediction during training
but omits future video generation at
inference~\citep{yuan2026fast}.
DreamZero instead uses caching to reuse intermediate results
and avoid repeated computation~\citep{ye2026world}.
These approaches reduce the amount of computation performed
during inference.
Our method addresses a complementary source of cost by
lowering the precision of weights and activations in the
retained operations.
It therefore complements computation reduction and caching
while preserving joint world--action denoising.

\section{Preliminary and Motivation}
\label{sec:motivation}
\label{sec:preliminary}

\subsection{WAM Quantization}
\label{sec:wam_inference}

WAMs jointly generate future world states and robot actions
through iterative denoising.
At each denoising step, a shared Transformer jointly processes
heterogeneous semantic streams, where each stream consists of
tokens encoding a particular type of information, such as video
frames, robot actions, value estimates, or proprioceptive states.
Video streams provide visual context for action generation,
while predicted actions drive the robot's interaction with
the environment.
Reducing the precision of linear operations in this shared
Transformer can lower the cost of joint denoising, but also
introduces errors into these coupled streams.

To characterize these errors, consider a linear operation at
a fixed layer $\ell$, denoising step $\tau$, and semantic stream
$s$, with input $\mathbf{X}$, weight $\mathbf{W}$, and
full-precision output $\mathbf{Y}$.
Using per-tensor quantization as an example, symmetric $k$-bit
quantization maps $\mathbf{X}$ to its integer representation as
\begin{equation}
\mathbf{Q}_{\mathbf{X}}
=
\operatorname{clip}
\left(
    \operatorname{round}
    \left(
        \frac{\mathbf{X}}{\Delta_{\mathbf{X}}}
    \right),
    -q_{\max}, q_{\max}
\right),
\qquad
\Delta_{\mathbf{X}}
=
\frac{c_{\mathbf{X}}}{q_{\max}},
\label{eq:wam_symmetric_quantization}
\end{equation}
where $c_{\mathbf{X}}>0$ is the clipping threshold,
$\Delta_{\mathbf{X}}$ is the quantization step size,
and $q_{\max}=2^{k-1}-1$.
Rounding and clipping are applied elementwise.
The weight $\mathbf{W}$ is quantized in the same manner using
its corresponding bit-width and step size, yielding
\begin{equation}
\mathbf{Y}
=
\mathbf{X}\mathbf{W}
\approx
\widehat{\mathbf{Y}}
=
\mathcal{Q}(\mathbf{X})\,\mathcal{Q}(\mathbf{W})
=
\Delta_{\mathbf{X}}\Delta_{\mathbf{W}} \cdot
\mathbf{Q}_{\mathbf{X}}\mathbf{Q}_{\mathbf{W}},
\label{eq:wam_quantized_linear}
\end{equation}
where $\mathcal{Q}$ denotes quantization followed by
dequantization.
Thus, weight and activation quantization jointly determine
the deviation of the layer output from the full-precision.

To compare these deviations across computational locations,
we explicitly index the same outputs as
$\mathbf{Y}_{\ell,\tau,s}$ and
$\widehat{\mathbf{Y}}_{\ell,\tau,s}$,
and define the normalized local reconstruction error as
\begin{equation}
\epsilon_{\ell,\tau,s}
=
\frac{
    \left\|
    \widehat{\mathbf{Y}}_{\ell,\tau,s}
    -
    \mathbf{Y}_{\ell,\tau,s}
    \right\|_{F}
}{
    \left\|
    \mathbf{Y}_{\ell,\tau,s}
    \right\|_{F}
}.
\label{eq:local_error}
\end{equation}
This quantity measures the relative magnitude of local errors. However, similar local errors may lead to different deviations in final actions as they propagate through the coupled computation.



\subsection{Observations}
\label{sec:local_vs_downstream}


\begin{figure}[!t]
    \centering
    \includegraphics[
        width=0.85\linewidth,
        trim={0.13in 3.55in 2.75in 0.225in},
        clip
    ]{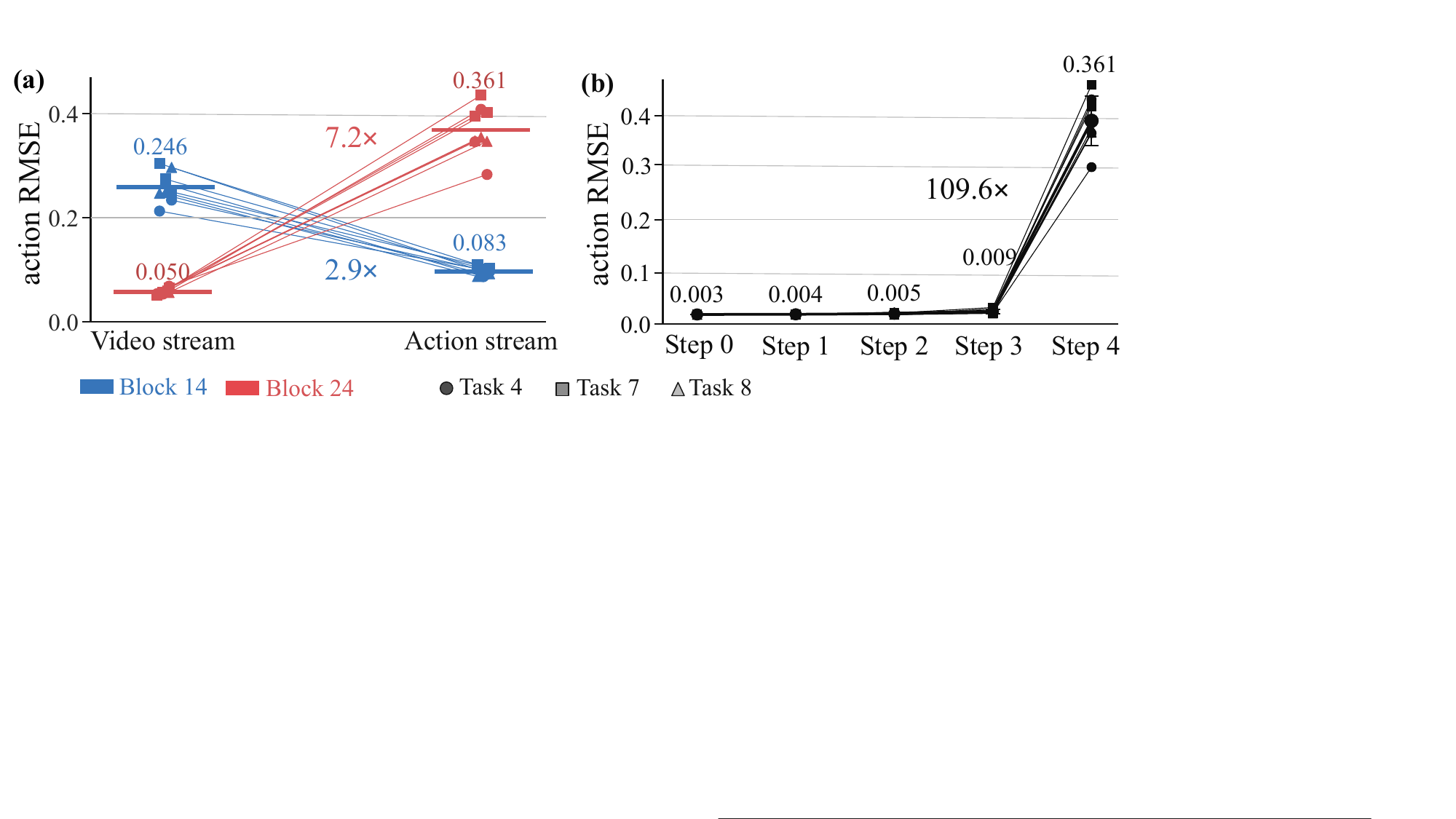}
    \caption{
    \textbf{Control effects of matched local errors.}
    \textbf{(a)} Video- and action-stream comparisons at
    Blocks~14 and~24 at the final denoising step, with local
    errors matched within each block.
    \textbf{(b)} Action-stream comparisons across Steps~0--4
    at Block~24, with local errors matched across steps.
    }
    \label{fig:matched_error}
    \vspace{-10pt}
\end{figure}
In this part, We examine how well local reconstruction error reflects
the effects of quantization on final actions and task success.
We inject rescaled W4A8 quantization residuals into selected
layer outputs of full-precision
Cosmos-Policy~\citep{kim2026cosmos}.
Within each comparison, we match the normalized local
reconstruction error in Eq.~\ref{eq:local_error}, while keeping
all remaining computation in full precision.
We measure standardized action RMSE relative to full-precision
predictions and closed-loop task success on three LIBERO tasks,
each with three initial states, yielding nine paired rollouts
per condition.
Appendix~\ref{app:preliminary_setup} provides the detailed protocol and complete results.

\textbf{The more influential stream varies across blocks.}
As shown in Fig.~\ref{fig:matched_error}(a), matched local
errors in the video and action streams have different
effects on final actions.
At the final denoising step, with local errors matched
within each block, video-stream errors at Block~14 produce
$2.9\times$ the action RMSE of action-stream errors.
At Block~24, this ordering reverses: action-stream errors
produce $7.2\times$ the RMSE of video-stream errors.
The difference also appears in closed-loop execution:
at Block~24, video-stream errors yield $9/9$ successful
rollouts, compared with $4/9$ for action-stream errors.
This observation shows that protection should be guided by
how strongly errors in each semantic stream affect final actions.

\textbf{Control impact changes across denoising steps.}
As shown in Fig.~\ref{fig:matched_error}(b), control impact
also varies when we hold the stream and block fixed and
inject matched local errors into the action stream at
Block~24 across Steps~0--4.
Step~4 produces $109.6\times$ the action RMSE of Step~0.
Task success is $4/9$ at Step~4, compared with $7/9$ or
$8/9$ at earlier steps.
This observation shows that a stream's action impact depends
on when the error occurs, motivating its characterization
across denoising steps as well as layers and streams.

These observations call for selective error reduction across
semantic streams and denoising steps based on their impact
on final actions.
However, each layer shares its weights across these streams
and steps, so quantization adjustments targeting one stream
at one step may also affect others.
The challenge is therefore to meet these differing protection
needs under shared weights and fixed 4-bit.
\vspace{-5pt}
\section{Methodology}
\label{sec:method}
\begin{figure*}[t]
    \centering
    \includegraphics[
        width=\textwidth,
        trim=0bp 0bp 0bp 0bp,
        clip
    ]{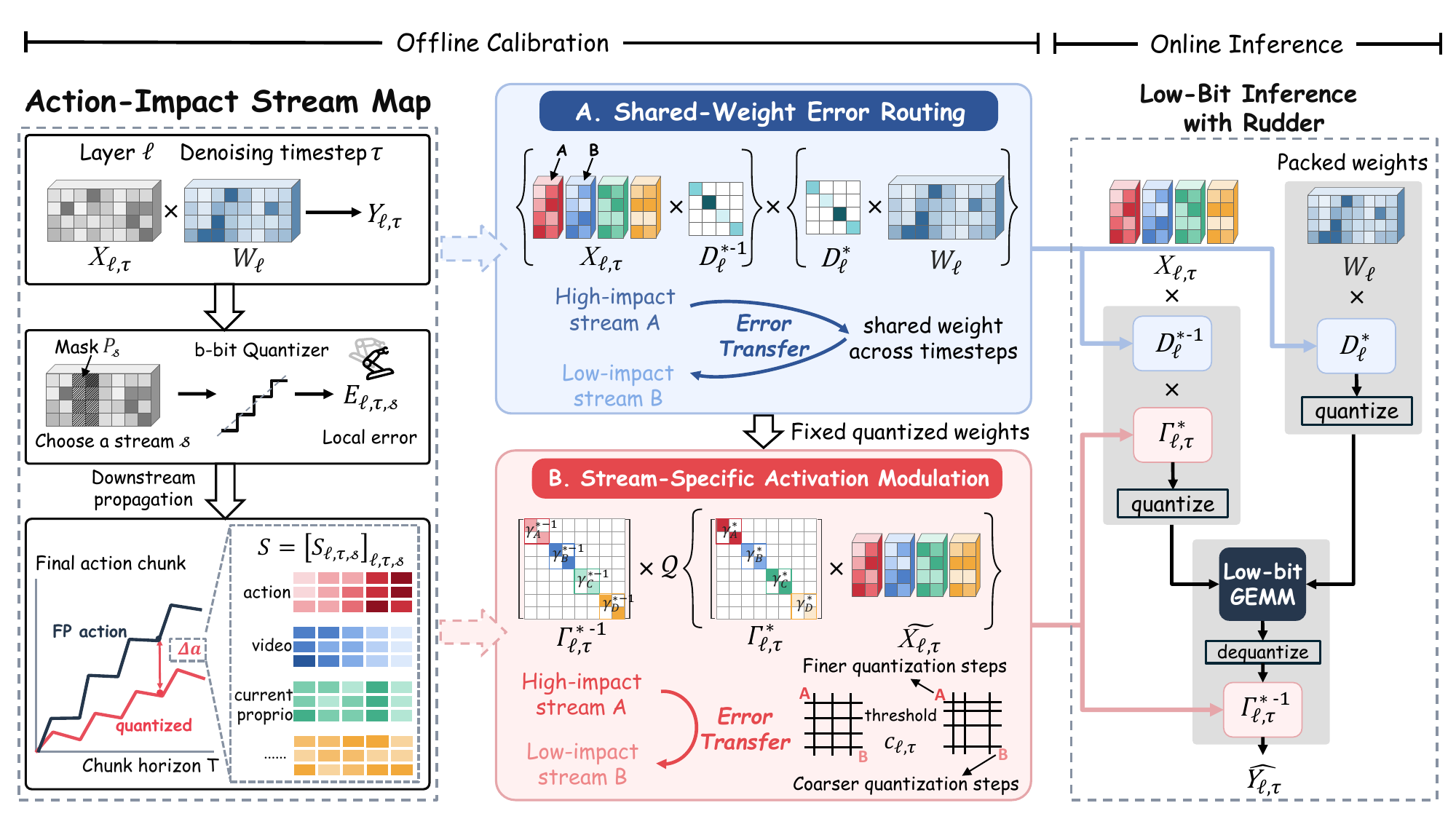}
    \caption{\textbf{Overview of \methodname{}.} $\Gamma_{\ell,\tau}$ is a diagonal matrix of token-wise stream gains. The upper-right gray box depicts offline weight preparation; the other gray boxes denote fused inference kernels.}
    \label{fig:overview}
    \vspace{-8pt}
\end{figure*}

\vspace{-5pt}
As illustrated in Figure~\ref{fig:overview}, \methodname{} first
constructs an \emph{Action-Impact Stream Map} that estimates how
local quantization errors affect final actions.
Guided by this map, \emph{Shared-Weight Error Routing} calibrates
shared channel scaling, followed by
\emph{Stream-Specific Activation Modulation}, which refines
activation quantization across streams and denoising steps
while keeping the quantized weights fixed.
The inference engine \emph{Rudder} fuses scaling and output
compensation into quantization and GEMM kernels for efficient
execution.


\subsection{Action-Impact Stream Map}
\label{sec:action_sensitivity_field}

To identify where quantization errors most affect final actions,
we score each layer--step--stream region by its impact on the
action output.
Before calibrating the scaling factors, we compute the local
quantization error at the target bit-widths:
\begin{equation}
\mathbf E_{\ell,\tau,s}
=
\widehat{\mathbf Y}_{\ell,\tau,s}
-
\mathbf Y_{\ell,\tau,s}.
\label{eq:stream_local_perturbation}
\end{equation}
Here, $\widehat{\mathbf Y}_{\ell,\tau,s}$ is the output for
stream $s$ after quantizing the weights and input activations
of linear layer $\ell$ at step $\tau$.
The input is taken from a full-precision forward pass
to exclude upstream quantization errors.

Injecting each stream's error and rerunning the remaining
full-precision computation requires
$\mathcal O(|\mathcal D|R)$ downstream forward evaluations
for $R$ regions and calibration set $\mathcal D$.
Instead, we use derivatives of the full-precision model
to estimate the resulting action changes.
Let $\mathbf a\in\mathbb R^m$ denote the vectorized final action
chunk, standardized using fixed, stabilized per-dimension
standard deviations.
Its Jacobian $\mathbf J_{\ell,\tau,s}$ with respect to
$\operatorname{vec}(\mathbf Y_{\ell,\tau,s})$ gives the
first-order action change
$\mathbf J_{\ell,\tau,s}
\operatorname{vec}(\mathbf E_{\ell,\tau,s})$.
We define the action-impact score as
\begin{equation}
S_{\ell,\tau,s}
=
\left[
\frac{1}{m}
\mathbb E_{\mathcal D}
\left[
\left\|
\mathbf J_{\ell,\tau,s}
\operatorname{vec}(\mathbf E_{\ell,\tau,s})
\right\|_2^2
\right]
\right]^{1/2}.
\label{eq:dataset_action_sensitivity}
\end{equation}
Here, $\mathbb E_{\mathcal D}$ averages over calibration samples.
The score measures the RMS first-order action displacement;
larger values indicate greater estimated impact on final actions.

Forming these Jacobians requires $m$ coordinate
vector--Jacobian products (VJPs) per sample and stores $m$
entries for every local output element.
We instead estimate the squared scores using $P$ Rademacher
projections~\citep{hutchinson1989stochastic}.
Each projected reverse pass provides VJPs at all captured
linear outputs, reusable across their stream error matrices.
This yields scores for all $R$ regions with
$\mathcal O(|\mathcal D|P)$ reverse passes, without storing
dense Jacobians.
The map remains fixed during subsequent calibration.
Appendix~\ref{app:map_estimation} derives the unbiased
squared-score estimator and analyzes its cost and approximation
error; Appendices~\ref{app:field_structure}
and~\ref{app:field_stability} examine the map's structure
and stability.

\subsection{Shared-Weight Error Routing}
\label{sec:static_redistribution}

Within each layer, semantic streams and denoising steps share
the same weights but differ in how their errors affect final
actions. A scaling that reduces reconstruction error in one
region may increase it in another.
We therefore use the Action-Impact Stream Map to choose a shared
channel scaling that prioritizes regions with greater action impact.

For each linear layer, a positive diagonal matrix $\mathbf D_\ell$
scales the weights and inversely scales the input activations:
\begin{equation}
\widehat{\mathbf Y}_{\ell,\tau,s}(\mathbf D_\ell)
=
\mathcal Q\!\left(
\mathbf X_{\ell,\tau,s}\mathbf D_\ell^{-1}
\right)
\mathcal Q\!\left(\mathbf D_\ell\mathbf W_\ell\right).
\label{eq:static_equivalent_transform}
\end{equation}
The two scalings cancel in full precision but change the ranges
of weights and activations seen by the quantizer, allowing us
to reshape quantization error at fixed bit-widths.
Using the same $\mathbf D_\ell$ across streams and denoising
steps retains one quantized weight matrix per layer.

We select this shared scaling by minimizing reconstruction
error weighted by the map-derived importance
$\omega^D_{\ell,\tau,s}$
(Appendix~\ref{app:field_aggregation}):
\begin{equation}
\mathbf D_\ell^*
=
\underset{\mathbf D_\ell}{\arg\min}
\sum_{\tau,s}
\frac{\pi_\tau\omega^D_{\ell,\tau,s}}{n_s}
\mathbb E_{\mathcal D}\!\left[
\left\|
\widehat{\mathbf Y}_{\ell,\tau,s}(\mathbf D_\ell)
-\mathbf Y_{\ell,\tau,s}
\right\|_F^2
\right].
\label{eq:static_calibration_objective}
\end{equation}
Here, $\pi_\tau$ is the normalized calibration frequency
of step $\tau$, and $n_s$ is the number of tokens in stream $s$,
so that $1/n_s$ averages the loss over stream tokens.
The map weights penalize errors more strongly in high-impact
regions, guiding the trade-off when regions favor different scalings.

After calibration, we fold $\mathbf D_\ell^*$ into the weights,
yielding $\widehat{\mathbf W}_\ell
=\mathcal Q(\mathbf D_\ell^*\mathbf W_\ell)$.
These weights are stored as packed integers with corresponding
quantization scales and remain fixed during subsequent
activation calibration and inference.

\subsection{Stream-Specific Activation Modulation}
\label{sec:dynamic_redistribution}

As discussed in Section~\ref{sec:local_vs_downstream}, the relative
action impact of semantic streams changes across denoising steps.
A shared channel scaling provides one compromise across all steps,
but cannot follow these changing priorities.
We therefore adjust activation quantization while keeping
the weights fixed.

At each layer and step, a shared clipping threshold establishes
a common reference range and base quantization scale.
Stream gains adjust each stream's effective range relative to
this reference, allowing different streams to receive greater
protection at different steps.
We bound both the threshold and gains to limit extreme range
choices that can cause excessive clipping or coarse quantization.

For the transformed activation
$\widetilde{\mathbf X}_{\ell,\tau,s}
=\mathbf X_{\ell,\tau,s}(\mathbf D_\ell^*)^{-1}$,
the modulated output is
\begin{equation}
\begin{aligned}
\widehat{\mathbf Y}_{\ell,\tau,s}
(\gamma_{\ell,\tau,s},c_{\ell,\tau})
=
\gamma_{\ell,\tau,s}^{-1}
\mathcal Q\!\left(
\gamma_{\ell,\tau,s}\widetilde{\mathbf X}_{\ell,\tau,s};
c_{\ell,\tau}
\right)
\widehat{\mathbf W}_\ell.
\end{aligned}
\label{eq:dynamic_quantized_output}
\end{equation}
The inverse gain compensates for the input scaling at the output.
Using map-derived weights $\omega^\gamma_{\ell,\tau,s}$,
we jointly calibrate the gain vector and shared threshold:
\begin{equation}
\begin{gathered}
(\boldsymbol\gamma_{\ell,\tau}^*,c_{\ell,\tau}^*)
=
\underset{\boldsymbol\gamma,\;c}{\arg\min}
\sum_s \frac{\omega^\gamma_{\ell,\tau,s}}{n_s}
\mathbb E_{\mathcal D}\!\left[
\left\|
\widehat{\mathbf Y}_{\ell,\tau,s}(\gamma_s,c)
-\widetilde{\mathbf X}_{\ell,\tau,s}
\widehat{\mathbf W}_\ell
\right\|_F^2
\right]
\\[1mm]
\text{s.t.}\quad
c_{\min}\le c\le c_{\max},
\qquad
\gamma_{\min}\le\gamma_s\le\gamma_{\max}
\quad \forall s,
\\
\sum_s n_s\log\gamma_s=0.
\end{gathered}
\label{eq:dynamic_calibration_objective}
\end{equation}
All streams share the same positive bounds.
When these constraints require a trade-off across streams,
the map weights favor lower reconstruction errors in those
with greater impact on final actions.
Appendix~\ref{app:field_aggregation} details the weight construction
and effective quantization ranges.
The calibrated gains and thresholds are stored in a
layer--step lookup table, preserving uniform bit-widths
without online optimization.

\begingroup
\textbf{Optimization.}
With map weights fixed, we optimize Routing (Eq.~\ref{eq:static_calibration_objective}) followed by Modulation (Eq.~\ref{eq:dynamic_calibration_objective}), using log-space Adam and straight-through gradients. Modulation keeps channel scales and quantized weights fixed. Appendix~\ref{app:optimization} provides implementation details.
\par\endgroup

\subsection{Rudder}
\label{sec:fused_kernel_design}

\begin{figure}[t]
    \centering
    \includegraphics[
        width=\linewidth,
        trim=0bp 30bp 0bp 45bp,
        clip
    ]{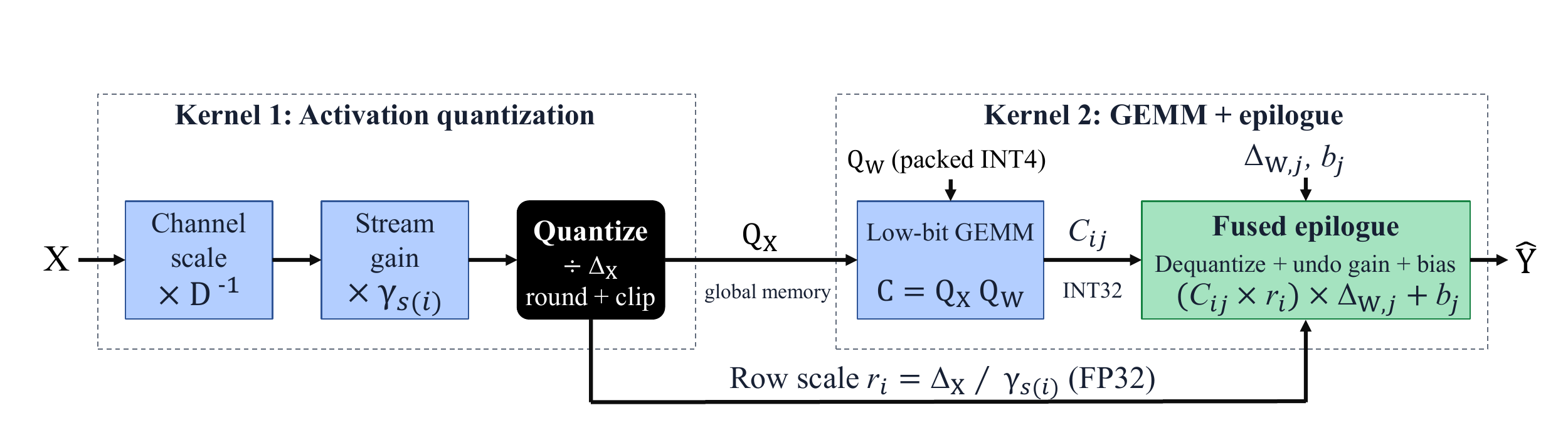}
    \caption{\textbf{Rudder's fused execution.}
    The first kernel combines channel scaling, stream gains,
    and activation quantization.
    The second performs low-bit GEMM and applies dequantization,
    inverse-gain compensation, and bias before writing the output.
    Only integer activations and row scales pass between kernels.
    Dashed boxes indicate kernel boundaries.}
    \label{fig:fused_kernel_design}
    \vspace{-8pt}
\end{figure}

Executing channel scaling, stream gains, and output compensation
separately adds kernel launches and intermediate memory traffic
at every denoising step.
Rudder, our 4-bit inference engine for WAMs, reduces this overhead
through two fused kernels while preserving stream-specific
quantization (Figure~\ref{fig:fused_kernel_design}).
Below, layer and step indices are omitted, and all scaling
parameters denote their calibrated values.

The first kernel applies channel scaling $\mathbf D^{-1}$
and stream gains during activation quantization.
It writes only integer activations $\mathbf Q_{\mathbf X}$
and row scales $r_i=\Delta_{\mathbf X}/\gamma_{s(i)}$,
where $s(i)$ identifies the stream of row $i$ and
$\Delta_{\mathbf X}=c/q_{\max}$.
The row scale incorporates inverse-gain compensation,
eliminating a separate compensation operation.

The second kernel reuses the packed INT4 weights
$\mathbf Q_{\mathbf W}$ across streams and steps.
Using CUTLASS~\citep{nvidia_cutlass}, it accumulates
$\mathbf C=\mathbf Q_{\mathbf X}\mathbf Q_{\mathbf W}$
in INT32 and produces the output in its epilogue:
\begin{equation}
\widehat Y_{ij}
=
\operatorname{cast}_{\mathrm{FP16/BF16}}
\!\left[
\bigl(\operatorname{FP32}(C_{ij})r_i\bigr)
\Delta_{\mathbf W,j}+b_j
\right].
\label{eq:kernel_epilogue}
\end{equation}
Here, $\Delta_{\mathbf W,j}$ is the weight scale for output
channel $j$, and $b_j$ is the optional bias.
This fusion avoids writing scaled floating-point activations
and INT32 accumulation results to global memory, retaining
only $\mathbf Q_{\mathbf X}$ and $\mathbf r$ between kernels.

{Rudder also fuses normalization and gate/residual operations and uses offline-calibrated thresholds. Appendix~\ref{app:execution_analysis} isolates the fusion gains with fixed parameters and GEMM work.}

\providecommand{\expTBD}[1]{\textit{[TODO: #1]}}
\providecommand{\expNA}{\textemdash}

\section{Experiments}
\label{sec:experiments}

\subsection{Setup}
\label{sec:experimental_setup}

\textbf{Models and benchmarks.}
We evaluate Cosmos-Policy~\citep{kim2026cosmos} and
FastWAMJoint~\citep{yuan2026fast} on LIBERO~\citep{liu2023libero},
and Cosmos-Edge~\citep{agarwal2026cosmos}
on RoboLab~\citep{yang2026robolab}. {We also evaluate Cosmos-Policy on RoboCasa tasks~\citep{nasiriany2024robocasa}.}

\textbf{Baselines and quantization settings.}
We evaluate W4A8 and W4A4 quantization, where W and A denote
the bit-widths of weights and activations, respectively.
For W4A8, we compare against
SmoothQuant~\citep{xiao2023smoothquant},
PTQ4DiT~\citep{wu2024ptq4dit}, and
Q-DiT~\citep{chen2025qdit}.
For W4A4, we compare against
QuaRot~\citep{ashkboos2024quarot},
Atom~\citep{zhao2024atom}, and
SVDQuant~\citep{li2025svdquant}.
The original full-precision (FP) model serves as a reference.
Within each model, all methods use the same checkpoint and
inference settings. {We additionally evaluate video-only activation quantization and QuantWAMs-FixedW4 on Cosmos-Policy / LIBERO (Appendix~\ref{app:additional_baselines}).}

\textbf{Evaluation metrics.}
We assess control performance using closed-loop success rate
(SR), supplemented by standardized action RMSE relative to FP
in ablation studies.
We assess efficiency using denoising latency and speedup
relative to FP, peak allocated GPU memory, and packed
weight storage.
The shared quantization scope and additional details are provided in Appendix~\ref{app:experimental_details}.

\subsection{Task Success}
\label{sec:task_success}

Table~\ref{tab:libero_main} compares task success
at FP, W4A8, and W4A4. We compare each quantized model with its own FP reference
and with baselines at the same nominal precision.

\begin{table}[htbp]
\centering

\caption{\textbf{Closed-loop success (\%).}
Mean $\pm$ sample SD across three evaluation seeds.
LIBERO entries average the four suites.
}
\label{tab:libero_main}
\label{tab:robolab_main}

\small
\setlength{\tabcolsep}{4pt}
\renewcommand{\arraystretch}{1.08}
\begin{tabular}{@{}llcccc@{}}
\toprule
& & \multicolumn{2}{c}{\textbf{LIBERO}}
& \textbf{RoboLab}
& \textbf{RoboCasa} \\
\cmidrule(lr){3-4}
\cmidrule(lr){5-5}
\cmidrule(lr){6-6}
Precision & Method
& Cosmos-Policy & FastWAMJoint
& Cosmos-Edge & Cosmos-Policy \\
\midrule
FP & Original
& $98.52\pm0.41$ & $98.90\pm0.39$
& $21.83\pm0.12$ & $63.33\pm0.72$ \\
\midrule
& SmoothQuant
& $97.93\pm0.32$ & $97.20\pm0.26$
& $13.67\pm0.25$ & $54.30\pm0.87$ \\
& PTQ4DiT
& $98.07\pm0.42$ & $97.60\pm0.25$
& $13.67\pm0.21$ & $58.61\pm1.05$ \\
& Q-DiT
& $97.66\pm0.09$ & $97.60\pm0.16$
& $12.17\pm0.36$ & $57.92\pm0.84$ \\
\rowcolor{gray!30}
\cellcolor{white}\multirow{-4}{*}{W4A8}
& SteerQuant
& $\mathbf{98.57}\pm0.25$ & $\mathbf{98.10}\pm0.22$
& $\mathbf{14.67}\pm0.30$ & $\mathbf{63.20}\pm2.55$ \\
\midrule
& QuaRot
& $97.93\pm0.23$ & $\mathbf{98.70}\pm0.21$
& $12.25\pm0.31$ & $54.59\pm2.60$ \\
& Atom
& $97.20\pm0.20$ & $98.20\pm0.27$
& $13.50\pm0.46$ & $53.05\pm0.87$ \\
& SVDQuant
& $98.13\pm0.23$ & $98.20\pm0.30$
& $16.00\pm0.27$ & $50.83\pm1.50$ \\
\rowcolor{gray!30}
\cellcolor{white}\multirow{-4}{*}{W4A4}
& SteerQuant
& $\mathbf{98.30}\pm0.26$ & $98.40\pm0.24$
& $\mathbf{17.33}\pm0.48$ & $\mathbf{60.14}\pm1.97$ \\
\bottomrule
\end{tabular}
\end{table}

\textbf{Results on LIBERO.}
{Across both models and precisions, \methodname{} maintains mean
closed-loop success within 0.8 percentage points (pp) of each model's
BF16 reference. On FastWAMJoint at W4A4, QuaRot reaches 98.70\%
versus 98.40\% for \methodname{} (Table~\ref{tab:libero_main}).}

\textbf{Results on RoboLab.}
On Cosmos-Edge, \methodname{} outperforms
all evaluated quantization baselines at both precision
settings (Table~\ref{tab:robolab_main}).
Under W4A8, it reaches 14.67\%, surpassing SmoothQuant
and PTQ4DiT by 1.00 pp.
Under W4A4, it attains 17.33\%, outperforming the strongest
baseline, SVDQuant, by 1.33 pp.
{All evaluated quantized methods fall below BF16 (21.83\%).
Appendix~\ref{app:robolab_calibration} discusses this gap and the
higher success rate observed at W4A4 than at W4A8.}

\begingroup
\textbf{Results on RoboCasa.} On 24 RoboCasa tasks with Cosmos-Policy, \methodname{} achieves
63.20\% at W4A8 and 60.14\% at W4A4, exceeding the strongest baselines by 4.59 and 5.55 pp. W4A8 remains close to BF16 (63.33\%) with a $1.526\times$ DiT speedup; W4A4 is 4.9\% faster than QuaRot in the main benchmark. Full efficiency results are in Table~\ref{tab:robocasa_full}.

\par\endgroup

\subsection{Inference Efficiency and Memory Footprint}
\label{sec:inference_efficiency}

Table~\ref{tab:efficiency} reports denoising latency,
speedup relative to FP, peak GPU memory, and weight storage.
All measurements use an NVIDIA RTX 4090 GPU, and all quantized configurations use low-bit kernels. Measurement details are provided in Appendix~\ref{app:experimental_details}.

\textbf{Denoising latency and speedup.}
{On an RTX 4090, W4A4 \methodname{} with Rudder achieves five-step DiT denoising speedups of $2.23\times$, $1.86\times$, and $1.69\times$ over BF16 on Cosmos-Policy, FastWAMJoint, and Cosmos-Edge, respectively.
Relative to the optimized QuaRot baseline, the corresponding latencies are 4.3\%, 12.5\%, and 8.5\% lower.
At W4A8, \methodname{} with Rudder also achieves lower latency than SmoothQuant on all three models.
These comparisons reflect the low-bit kernels and operator fusions used by each quantized implementation.}

\textbf{Memory and storage footprint.}
Peak allocated GPU memory includes resident weights,
activations, and temporary buffers, whereas packed weight
storage covers stored model weights only.
Table~\ref{tab:efficiency} reports these quantities
in GB and GiB, respectively.
Across models and precision settings, \methodname{}
reduces peak GPU memory by 24.7--72.2\% and packed weight
storage by 31.3--73.4\% relative to FP.

\begin{table*}[t]
\centering
\caption{\textbf{Inference efficiency and memory footprint
across models.}
Lat.: denoising latency (ms); Spd.: speedup relative to FP;
Peak: peak allocated GPU memory (GB);
Storage: packed weight storage (GiB). {The QuaRot row uses the optimized implementation identified in the second row of Table~\ref{tab:quarot_control}, with Hadamard--quantization fusion and output dequantization in the GEMM epilogue.}
PTQ4DiT and Q-DiT are omitted because their official
implementations do not provide low-bit inference kernels
for the evaluated settings.}
\label{tab:efficiency}
\label{tab:efficiency_memory}
\vspace{-5pt}
\scriptsize
\setlength{\tabcolsep}{1.9pt}
\renewcommand{\arraystretch}{1.08}
\begin{tabular}{@{}ll*{12}{c}@{}}
\toprule
\multicolumn{2}{c}{}
& \multicolumn{4}{c}{\textbf{Cosmos-Policy}}
& \multicolumn{4}{c}{\textbf{FastWAMJoint}}
& \multicolumn{4}{c}{\textbf{Cosmos-Edge}} \\
\cmidrule(lr){3-6}
\cmidrule(lr){7-10}
\cmidrule(lr){11-14}
Precision & Method
& Lat. $\downarrow$ & Spd. $\uparrow$
& Peak $\downarrow$ & Storage $\downarrow$
& Lat. $\downarrow$ & Spd. $\uparrow$
& Peak $\downarrow$ & Storage $\downarrow$
& Lat. $\downarrow$ & Spd. $\uparrow$
& Peak $\downarrow$ & Storage $\downarrow$ \\
\midrule
FP & Original
& 261.460 & $1.000\times$ & 4.241 & 3.644
& 360.450 & $1.000\times$ & 12.116 & 11.230
& 881.637 & $1.000\times$ & 8.492 & 6.276 \\
\midrule
& SmoothQuant
& 134.967 & $1.937\times$ & 1.607 & 1.191
& 263.368 & $1.369\times$ & 3.363 & 2.983
& 570.893 & $1.544\times$ & 6.405 & 4.314 \\
\rowcolor{gray!30}
\cellcolor{white}\multirow{-2}{*}{W4A8}
& \methodname{}
& 133.455 & $1.959\times$ & 1.642 & 1.191
& 260.977 & $1.381\times$ & 3.363 & 2.983
& 563.858 & $1.564\times$ & 6.397 & 4.312 \\
\midrule
\multirow{4}{*}{W4A4}
& QuaRot
& 122.657 & $2.132\times$ & 1.604 & 1.189
& 221.647 & $1.626\times$ & 3.363 & 2.983
& 569.943 & $1.547\times$ & 6.396 & 4.312 \\
& Atom
& 192.997 & $1.355\times$ & 1.679 & 1.258
& 309.021 & $1.166\times$ & 3.510 & 3.188
& 766.027 & $1.151\times$ & 6.463 & 4.361 \\
& SVDQuant
& 131.093 & $1.994\times$ & 1.755 & 1.321
& 348.767 & $1.033\times$ & 3.756 & 3.371
& 581.278 & $1.517\times$ & 6.503 & 4.411 \\
\rowcolor{gray!30}
\cellcolor{white} & \methodname{}
& 117.322 & $2.229\times$ & 1.645 & 1.194
& 193.916 & $1.859\times$ & 3.370 & 2.983
& 521.589 & $1.690\times$ & 6.398 & 4.314 \\
\bottomrule
\end{tabular}
\end{table*}

\begingroup
\textbf{Execution and calibration costs.}
With calibrated parameters and GEMM work fixed, fusing diagonal scaling into W4A8 activation quantization yields a $1.36\times$ speedup. Block-level fusion adds a further $1.08\times$ (Table~\ref{tab:rudder_execution}).
Separate roofline and overhead analyses are in
Appendix~\ref{app:execution_analysis}.
SteerQuant also calibrates faster than several baselines;
detailed timing comparisons are provided in
Appendix~\ref{app:calibration_cost}.
\par\endgroup

\subsection{Ablation Studies}
\label{sec:ablations}

\begin{table}[t]
\centering
\caption{\textbf{Contribution of individual components
and action guidance.}
Standardized action RMSE is averaged over W4A8 and W4A4 on Cosmos-Policy,
using 1,800 observations from 600 trajectories across 40 tasks.
Paired, unadjusted 95\% CIs are computed by resampling
whole trajectories within each task--seed stratum and refer to
$\Delta\mathrm{RMSE}
= \mathrm{RMSE}_{\mathrm{\methodname{}}}-\mathrm{RMSE}_{\mathrm{row}}$. Negative values favor \methodname{}. {SR averages W4A8 and W4A4; each precision uses 40 tasks, three seeds and five rollouts per task per seed (600 rollouts).}}
\label{tab:ablation_components}
\vspace{-5pt}
\footnotesize
\setlength{\tabcolsep}{3pt}
\renewcommand{\arraystretch}{1.08}
\begin{tabular}{@{}lcccccc@{}}
\toprule
Variant & Routing & Modulation & Weighting
& RMSE $\downarrow$ & 95\% CI of $\Delta$RMSE & {SR (\%) $\uparrow$} \\
\midrule
Base
& $\times$ & $\times$ & Uniform
& 0.1581 & $[-0.0430,\,-0.0357]$ & {2.17} \\
Routing only
& $\checkmark$ & $\times$ & Map-derived
& 0.1224 & $[-0.0044,\,-0.0026]$ & {78.16} \\
Modulation only
& $\times$ & $\checkmark$ & Map-derived
& 0.1529 & $[-0.0368,\,-0.0312]$ & {6.50} \\
Both, uniform
& $\checkmark$ & $\checkmark$ & Uniform
& 0.1416 & $[-0.0262,\,-0.0052]$ & {17.83} \\
\rowcolor{gray!30}
\methodname{}
& $\checkmark$ & $\checkmark$ & Map-derived
& \textbf{0.1189} & -- & {97.33} \\
\bottomrule
\end{tabular}
\end{table}

\textbf{Contribution of individual components and action guidance.}
Table~\ref{tab:ablation_components} evaluates
Shared-Weight Error Routing, Stream-Specific Activation
Modulation, and weighting derived from the
Action-Impact Stream Map.
Shared-Weight Error Routing alone achieves lower action RMSE
than Stream-Specific Activation Modulation alone, and combining
both components further improves action fidelity.
The full method reduces RMSE by 24.79\% relative to the base
quantizer.
{Adding Modulation to Routing also increases mean SR
from 78.16\% to 97.33\%, demonstrating its contribution
to closed-loop task performance.}
Replacing map-derived weighting with uniform weighting
increases RMSE {and reduces mean SR to 17.83\%},
supporting the contribution of guidance
from the Action-Impact Stream Map.

\begin{figure}[t]
    \centering
    \begin{minipage}[t]{0.49\textwidth}
        \vspace{0pt}
        \centering
        \includegraphics[
            width=\linewidth,
            trim=0bp 20bp 0bp 50bp,
            clip
        ]{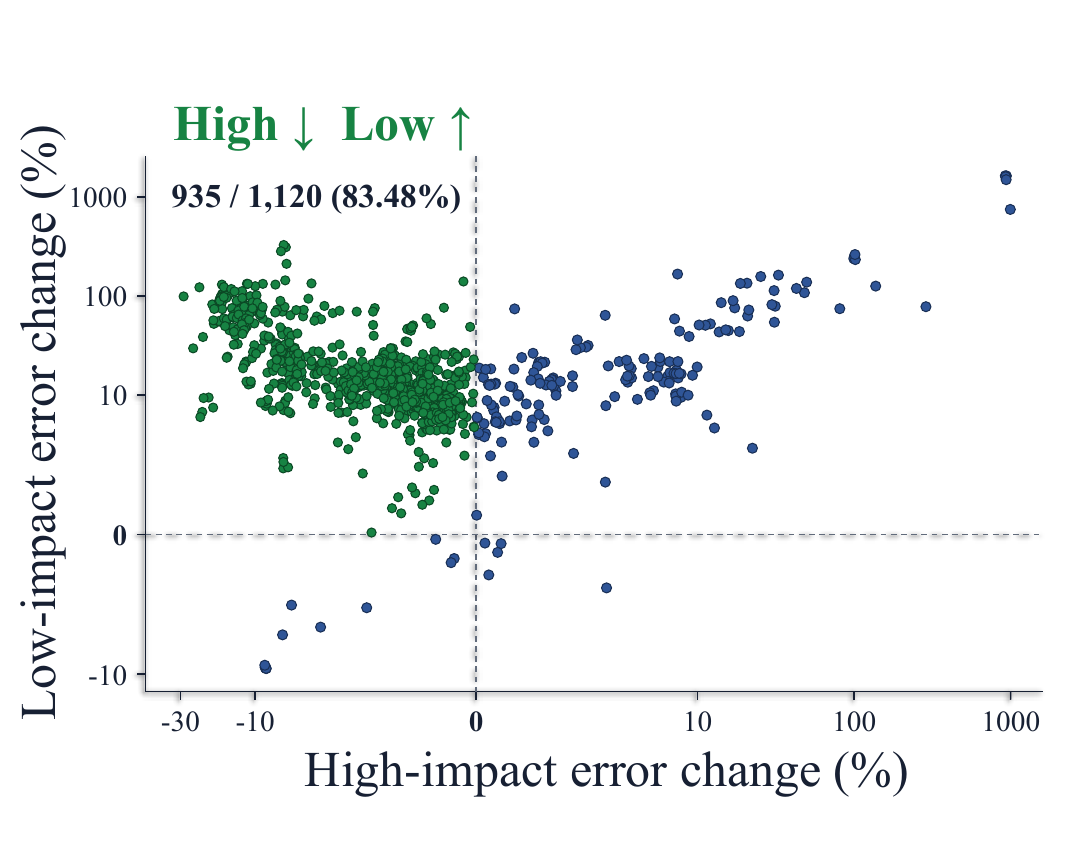}
        \par\smallskip
    \end{minipage}\hfill
    \begin{minipage}[t]{0.47\textwidth}
        \vspace{0pt}
        \centering
        \includegraphics[width=\linewidth]{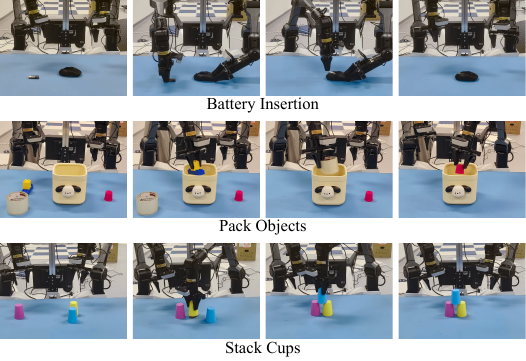}
        \par\smallskip
    \end{minipage}
    \caption{\textbf{Action-guided quantization error analysis
    and real-world execution.}
    \textbf{Left:} Cosmos-Policy W4A8 on 1,800 observations.
Errors are token-normalized squared Frobenius differences from FP outputs
under matched FP inputs, averaged over observations and streams within each group.
Axes show percentage changes relative to uniform weighting on symmetric logarithmic scales.
Green points indicate lower high-impact and higher low-impact error.
\textbf{Right:} Representative \methodname{} executions on the
AgileX Cobot Magic dual-arm platform. Rows show battery
insertion, object packing, and cup stacking, with time
progressing from left to right.}
    \label{fig:error_redistribution}
    \vspace{-8pt}
\end{figure}

\textbf{Quantization error analysis.}
Figure~\ref{fig:error_redistribution} (left) compares local output
errors under uniform weighting and weighting derived from
the Action-Impact Stream Map.
Within each layer--step unit, we rank streams by their
reference action-impact scores estimated on the calibration set.
The top and bottom three streams form the high- and
low-impact groups, respectively, with group assignments
fixed across variants.
In 935 of the 1,120 units (83.48\%), map-derived weighting
reduces error in the high-impact group while increasing
it in the low-impact group.
This pattern supports preferential error reduction in
streams with greater estimated impact on final actions,
while allowing larger errors in lower-impact streams.


\begingroup
\subsection{Comparison with Additional Baselines}
\label{sec:additional_baselines}

Table~\ref{tab:additional_baselines} compares additional
baselines on Cosmos-Policy / LIBERO, all using shared
INT4 target weights. Video-only retains BF16 non-video
activations; QuantWAMs-FixedW4 retains mixed activation
protections averaging 5.65 bits.

SteerQuant achieves higher SR and lower latency than both
baselines. Notably, video-only quantization collapses to
12.07\% SR at W4A4, despite retaining BF16 non-video activations.
Against QuantWAMs-FixedW4, SteerQuant improves SR
by 2.70 pp while reducing latency by 30.71\% and peak
memory by 40.16\%.
Implementation details and performance analyses are in
Appendix~\ref{app:additional_baselines}.
\par\endgroup

\begin{table}[htbp]
\centering

\caption{\textbf{Additional baselines on Cosmos-Policy / LIBERO.}
SR is mean $\pm$ sample SD across three seeds.
Asterisks denote mixed-activation configurations described
in the text. Speedup is relative to BF16.}
\label{tab:additional_baselines}

\footnotesize
\setlength{\tabcolsep}{3pt}
\renewcommand{\arraystretch}{1.08}
\begin{tabular}{@{}llccccc@{}}
\toprule
Precision & Method
& SR (\%) $\uparrow$
& \shortstack{Latency\\(ms) $\downarrow$}
& Speedup $\uparrow$
& \shortstack{Peak memory\\(GB) $\downarrow$}
& \shortstack{Storage\\(GiB) $\downarrow$} \\
\midrule
FP & Original
& $98.52\pm0.41$
& 261.460 & $1.000\times$ & 4.241 & 3.644 \\
\midrule
W4A8* & Video-only
& $97.80\pm0.53$
& 316.597 & $0.826\times$ & 1.639 & 1.186 \\
\rowcolor{gray!30}
W4A8 & SteerQuant
& $\mathbf{98.57}\pm0.25$
& \textbf{133.455} & $\mathbf{1.959}\times$
& 1.642 & 1.191 \\
\midrule
W4A4* & Video-only
& $12.07\pm0.60$
& 284.523 & $0.919\times$ & 1.639 & 1.186 \\
W4A4* & QuantWAMs-FixedW4
& $95.60\pm0.56$
& 169.320 & $1.544\times$ & 2.749 & 1.400 \\
\rowcolor{gray!30}
W4A4 & SteerQuant
& $\mathbf{98.30}\pm0.26$
& \textbf{117.322} & $\mathbf{2.229}\times$
& 1.645 & 1.194 \\
\bottomrule
\end{tabular}
\end{table}

\subsection{Performance on Real-world Tasks}
We evaluate FastWAMJoint fine-tuned on three real-world tasks
using the AgileX Cobot Magic dual-arm platform and an NVIDIA
RTX 4090 GPU (24 GB). Both methods use the same checkpoint and initial conditions, with 12 trials per task.
Figure~\ref{fig:error_redistribution} (right) shows representative executions.
As shown in Table~\ref{tab:steerquant_real_robot}, W4A8 \methodname{} achieves
82.41\% average task score versus 75.93\% for BF16, with a $1.35\times$ end-to-end
inference speedup. Although battery insertion score decreases, the overall
results show faster inference with higher average task success.

\begin{table}[t]
\centering
\caption{\textbf{Real-world task performance and inference efficiency.}
Task scores are reported in percent and averaged equally across tasks. Scoring details are provided in Appendix~\ref{app:experimental_details}. We report the median end-to-end inference latency and speedup, measured from sending a policy request to receiving the predicted actions.
Robot motion execution time is excluded.}
\label{tab:steerquant_real_robot}
\small
\vspace{-5pt}
\setlength{\tabcolsep}{4pt}
\renewcommand{\arraystretch}{1.1}
\begin{tabularx}{\linewidth}{l*{6}{>{\centering\arraybackslash}X}}
\toprule
Method
& \shortstack{Battery\\Insertion}
& \shortstack{Pack\\Objects}
& \shortstack{Stack\\Cups}
& \shortstack{Avg. score\\(\%) $\uparrow$}
& \shortstack{Latency\\(ms) $\downarrow$}
& \shortstack{Speedup\\$\uparrow$} \\
\midrule
FP (BF16)
& 83.33 & 69.44 & 75.00 & 75.93 & 462.79 & $1.00\times$ \\
\rowcolor[gray]{0.85}
SteerQuant (W4A8)
& 75.00 & 88.89 & 83.33 & 82.41 & 341.97 & $1.35\times$ \\
\bottomrule
\end{tabularx}
\end{table}
\section{Limitations and Future Work}
Future work can broaden the evaluation and deployment of SteerQuant in several directions.
First, our efficiency evaluation is conducted on an NVIDIA RTX 4090. Extending Rudder to other GPU architectures and resource-constrained edge devices~\citep{wang2025d2moe} would broaden its deployment opportunities.
Second, our evaluation covers three WAMs.
{SteerQuant requires action outputs to construct its
Action-Impact Stream Map; extending it to frame-only models~\citep{miao2026onlinewm}
requires a different sensitivity target.}
Future work could explore additional model architectures and scales to further assess the applicability of action-guided quantization. 
Third, our real-world experiments focus on three manipulation tasks using a dual-arm robot. Expanding evaluation to more tasks and robot platforms would provide a broader picture of SteerQuant’s practical benefits.

\section{Conclusion}
We presented SteerQuant, a post-training quantization framework that guides error allocation across heterogeneous semantic streams according to their influence on final actions. Its action-guided scaling coordinates shared weights and stream-specific activation ranges across denoising steps, preserving one static quantized weight matrix per layer at fixed bit-widths. The Rudder inference engine integrates these transformations into fused low-bit kernels. Across three WAMs, SteerQuant preserves near-BF16 success on LIBERO, outperforms the evaluated same-precision baselines on RoboLab, {improves RoboCasa success with Cosmos-Policy over the strongest evaluated quantization baseline by up to 5.55 percentage points,} and achieves up to 2.23× denoising speedup with reduced GPU memory usage. Real-world W4A8 deployment delivers a 1.35× end-to-end inference speedup while maintaining average task success. These findings support downstream action impact as a useful guide for efficient quantization of coupled world–action computation.

\ifarxiv
\else
  \newpage
  \subsection*{AI use statement}
We used generative AI tools to polish the language and improve the clarity and readability of the manuscript. All AI-assisted revisions were reviewed and verified by the authors, who take full responsibility for the content, claims, and artifacts presented in this work.

\subsection*{Ethics statement}
This work studies post-training quantization of pretrained world–action models. We use LIBERO and DROID data for calibration and evaluate on {LIBERO, RoboLab, and RoboCasa}, in accordance with their respective terms of use. Real-world experiments involve three manipulation tasks on the AgileX Cobot Magic dual-arm platform. Quantization can alter policy behavior and task performance, and improved average success rates do not guarantee safe execution in individual cases. Our evaluations do not establish safety for deployment around humans or in safety-critical settings.

\subsection*{Reproducibility statement}
Sections 4 and 5 describe the proposed quantization method, the Rudder inference engine, and the experimental setup. Appendix A details the controlled intervention experiments; Appendices B and C provide the action-impact estimation procedure, supporting analysis, and calibration formulations; and Appendix D specifies the quantization scope, calibration data, and evaluation protocols. {Appendix~\ref{app:revision_experiments} reports additional benchmarks, baseline adaptations, calibration cost and sensitivity; Appendix~\ref{app:execution_analysis} provides controlled execution and roofline analyses.} We plan to release the calibration and evaluation code, together with the Rudder inference kernels, upon publication.

\fi



\bibliography{iclr2027_conference}
\bibliographystyle{iclr2027_conference}

\appendix

\section{Experimental Details for Preliminary Analysis}
\label{app:preliminary_setup}

\subsection{Experimental Setup}

\begin{figure}[h]
    \centering
    \includegraphics[
        width=0.82\linewidth,
        trim={20bp 20bp 340bp 4bp},
        clip
    ]{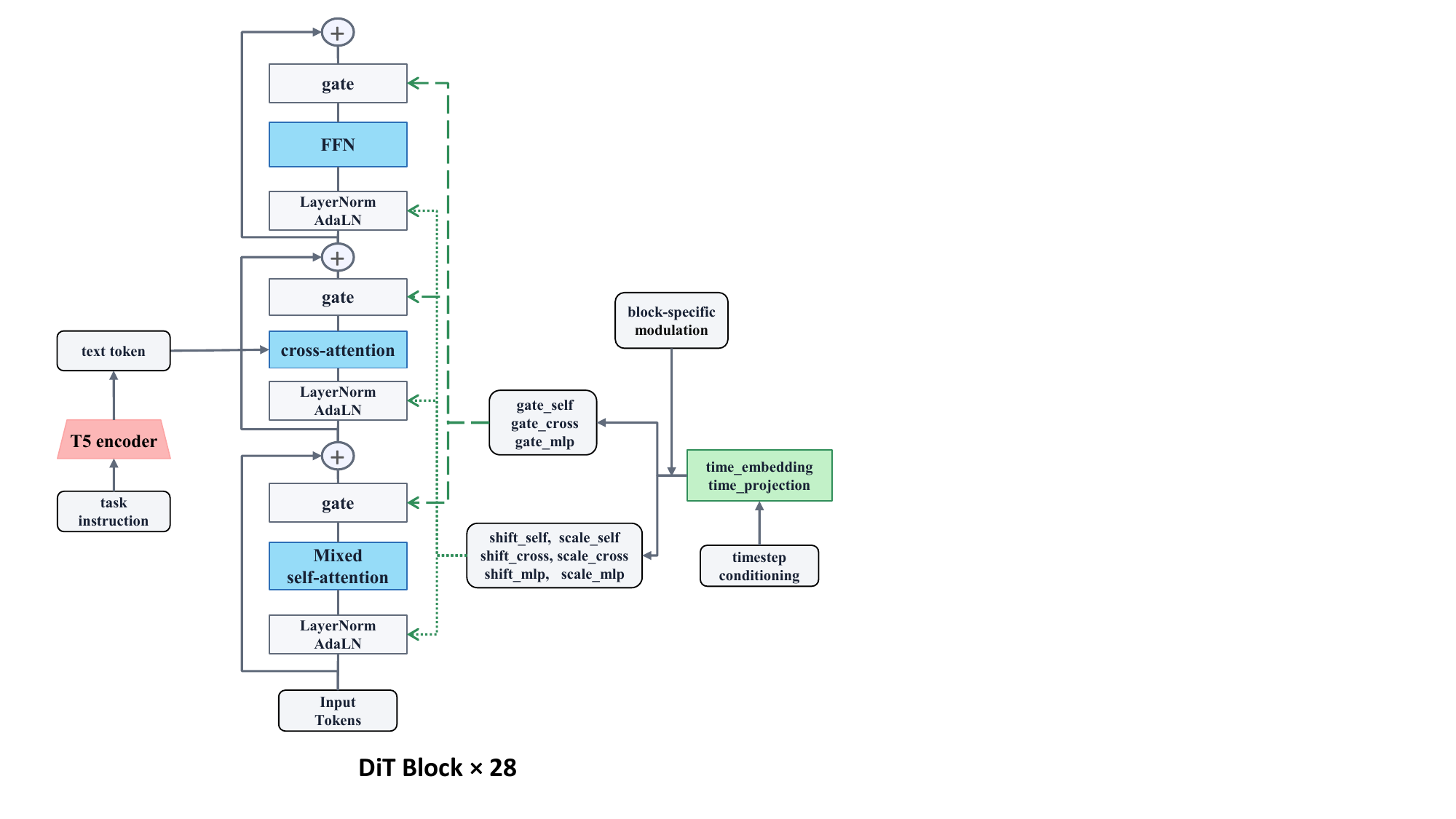}
    \caption{\textbf{DiT backbone of the evaluated Cosmos-Policy
    checkpoint.} The Cosmos-Policy checkpoint inherits the 28-block Cosmos-Predict2-2B Video2World
    diffusion Transformer without architectural modifications~\citep{kim2026cosmos}.
    Each block applies AdaLN-modulated mixed self-attention,
    T5-conditioned cross-attention, and a feedforward network. The three blue modules contain ten Linear layers; each residual intervention targets one Linear output.}
    \label{fig:appendix_cosmos_policy_dit}
    \vspace{-8pt}
\end{figure}
We evaluated the full-precision
Cosmos-Policy checkpoint on LIBERO-10
Tasks~4, 7, and~8. These tasks require placing two mugs on separate
plates, placing two packaged objects in a basket, and placing two moka
pots on a stove, respectively. We used three default initial states per
task, yielding nine paired rollouts per condition. All evaluations used
policy seed 195, deterministic environment resets with seed 0, five
denoising steps, and action chunks of shape \(16\times7\).

Figure~\ref{fig:appendix_cosmos_policy_dit} summarizes the Transformer block used by the evaluated checkpoint. Its mixed self-attention, cross-attention, and feedforward modules contain ten Linear layers. Each intervention targets the output of a single Linear layer; we identify its location by the Transformer block and focus on block depth rather than comparisons among Linear families. The model contains 28 Transformer blocks, indexed from 1, whereas denoising steps are indexed from 0. The stream–depth experiment intervened at Step 4 on the video or action stream at Blocks 14 and 24. The step-wise experiment fixed Block 24 and the action stream while varying the intervention over Steps 0–4.

\subsection{Matched Quantization-Residual Intervention}

For each policy query, we first run an FP pass and record the input and
output \(\mathbf{Y}^{\mathrm{FP}}_{\ell,\tau}\) of the target Linear layer
\(\ell\) at denoising step \(\tau\). Using the same FP input, we quantize
the layer's weights and input activations with W4A8 round-to-nearest
quantization and compute the resulting output
\(\mathbf{Y}^{\mathrm{W4A8}}_{\ell,\tau}\). This produces the local
quantization residual
\begin{equation}
\mathbf{r}_{\ell,\tau}
=
\mathbf{Y}^{\mathrm{W4A8}}_{\ell,\tau}
-
\mathbf{Y}^{\mathrm{FP}}_{\ell,\tau}.
\label{eq:appendix_native_residual}
\end{equation}
Let \(P_s\) extract the output rows corresponding to stream \(s\), so that
\(\mathbf{Y}^{\mathrm{FP}}_{\ell,\tau,s}
= P_s(\mathbf{Y}^{\mathrm{FP}}_{\ell,\tau})\).
At the injection site, we retain all unselected streams at their FP
values and scale the selected residual independently for every policy query:
\begin{equation}
\begin{gathered}
\widetilde{\mathbf{Y}}_{\ell,\tau,s}
=
\mathbf{Y}^{\mathrm{FP}}_{\ell,\tau,s}
+
\alpha_{\ell,\tau,s} P_s(\mathbf{r}_{\ell,\tau}),
\\
\alpha_{\ell,\tau,s}
=
\epsilon_{\ell,\tau,s}
\frac{
\left\|\mathbf{Y}^{\mathrm{FP}}_{\ell,\tau,s}\right\|_F
}{
\left\|P_s(\mathbf{r}_{\ell,\tau})\right\|_F
}.
\end{gathered}
\label{eq:appendix_matched_residual}
\end{equation}
The remaining downstream computation is performed in full precision.
Here, \(\epsilon_{\ell,\tau,s}\) is the target relative local error.
Consequently,
\[
\frac{
\left\|
\widetilde{\mathbf{Y}}_{\ell,\tau,s}
-
\mathbf{Y}^{\mathrm{FP}}_{\ell,\tau,s}
\right\|_F
}{
\left\|\mathbf{Y}^{\mathrm{FP}}_{\ell,\tau,s}\right\|_F
}
\approx \epsilon_{\ell,\tau,s}.
\]
We used \(\epsilon_{\ell,\tau,s}=0.25\) at Block~14 and
\(\epsilon_{\ell,\tau,s}=0.50\) at Block~24.
For the step-wise comparison, \(\epsilon_{\ell,\tau,s}=0.50\)
was fixed across all five steps.

The FP and intervened action chunks were predicted from the same
observation and sampler seed. For rollout \(r\) and policy query \(q\),
the standardized action RMSE is
\begin{equation}
e_{r,q}
=
\sqrt{
\frac{1}{16\times7}
\sum_{k=1}^{16}
\sum_{d=1}^{7}
\left(
\frac{
\widetilde{a}_{r,q,k,d} - a^{\mathrm{FP}}_{r,q,k,d}
}{
\sigma_d+10^{-6}
}
\right)^2
},
\label{eq:appendix_action_rmse}
\end{equation}
where \(\sigma_d\) is the action standard deviation from the LIBERO
training statistics. We first average \(e_{r,q}\) over all policy queries
in an episode, and then report the mean and sample standard deviation
over the nine paired episodes. The intervened actions, rather than the
auxiliary FP predictions, are executed in the closed-loop environment.

\begin{table*}[t]
\centering
\setlength{\belowcaptionskip}{5pt}
\caption{\textbf{Complete results of the controlled intervention
experiments.} Action RMSE is reported as mean $\pm$ s.d.\ over
nine task--initial-state rollouts. Success rate is reported as the
number of successful rollouts out of nine. Block indices are one-based
and step indices are zero-based.}
\label{tab:preliminary_complete_results}

\small
\vspace{-5pt}
\setlength{\tabcolsep}{3.7pt}
\begin{tabular}{@{}lccccccc@{}}
\toprule
Setting & Block & Step & Stream
& Target error & Observed error
& Action RMSE & Success Rate \\
\midrule
FP
& -- & -- & -- & -- & --
& -- & \(7/9\) \\
\midrule
Stream $\times$ depth
& 14 & 4 & Video
& 0.25 & 0.250007
& \(0.245637\pm0.029194\) & \(5/9\) \\

Stream $\times$ depth
& 14 & 4 & Action
& 0.25 & 0.250016
& \(0.083329\pm0.007876\) & \(9/9\) \\

Stream $\times$ depth
& 24 & 4 & Video
& 0.50 & 0.499968
& \(0.050042\pm0.005754\) & \(9/9\) \\

Stream $\times$ depth
& 24 & 4 & Action
& 0.50 & 0.500003
& \(0.361126\pm0.045779\) & \(4/9\) \\

\midrule
Step
& 24 & 0 & Action
& 0.50 & 0.499986
& \(0.003295\pm0.000331\) & \(7/9\) \\

Step
& 24 & 1 & Action
& 0.50 & 0.500002
& \(0.003606\pm0.000432\) & \(8/9\) \\

Step
& 24 & 2 & Action
& 0.50 & 0.500001
& \(0.004670\pm0.001770\) & \(7/9\) \\

Step
& 24 & 3 & Action
& 0.50 & 0.500004
& \(0.008696\pm0.003911\) & \(8/9\) \\

Step
& 24 & 4 & Action
& 0.50 & 0.500003
& \(0.361126\pm0.045779\) & \(4/9\) \\
\bottomrule
\end{tabular}
\end{table*}

\subsection{Additional Statistical Details}

Table~\ref{tab:preliminary_complete_results} provides the complete
numerical results underlying Fig.~\ref{fig:matched_error}, including the
realized local errors, condition-level action RMSE summaries, and
closed-loop success counts.

For the five repeated step conditions, a Friedman test detected a
difference in action deviation
($\chi^2(4)=30.22$, $P=4.41\times10^{-6}$), indicating that matched
local errors did not produce equal action deviations across
denoising steps. The mean Step~4--Step~0 difference in action RMSE
was $0.3578$, with a task-aware paired-bootstrap 95\% confidence
interval of $0.3214$--$0.3987$. This interval excludes zero,
supporting a larger action deviation from intervention at
Step~4 than at Step~0 under matched local errors.

\section{Action-Impact Stream Map: Construction and Analysis}
\label{app:w4a8_action_sensitivity}

\subsection{Calibration Setup of Cosmos-Policy}
\label{app:field_setup}

We construct a separate Action-Impact Stream Map for each target weight–activation bit-width configuration. The construction details and the structure and stability analyses presented in this appendix focus on the W4A8 map for Cosmos-Policy, computed using 320 LIBERO calibration observations.
Table~\ref{tab:w4a8_field_configuration} summarizes the setup.

\textbf{Region definition.}
Each region is indexed by a Linear layer $\ell$,
a denoising step $\tau$, and a semantic stream $s$.
The index $\ell$ identifies both the Transformer block
and the Linear family within that block.
Each of the 28 blocks contains ten analyzed Linear layers:
the query, key, value, and output projections of
self-attention and cross-attention, and two MLP projections. Eight of these Linear families operate on nine temporal
semantic streams: blank control, current proprioception,
current wrist image, current primary image, action,
future proprioception, future wrist image,
future primary image, and value.
For these families, the stream mask $P_s$ selects the
output rows corresponding to stream $s$.
The remaining two families, the cross-attention key and
value projections, operate on global text context.
Each is treated as a single region with $P_s=I$.
Across five denoising steps, this yields
$28\times5\times(8\times9+2)=10{,}360$ regions per observation.

\textbf{Map estimation.}
We estimate the action response to the joint
weight--activation quantization error using
16 Rademacher projections and vector--Jacobian products
along the full-precision trajectory.
For each calibration observation, all map-estimation passes use the same conditioning and diffusion random draws.

\begin{table}[t]
\centering
\setlength{\belowcaptionskip}{5pt}
\caption{Calibration setup for constructing the
Cosmos-Policy Action-Impact Stream Map.}
\label{tab:w4a8_field_configuration}
\small
\vspace{-5pt}
\begin{tabular}{@{}ll@{}}
\toprule
Property & Configuration \\
\midrule
Calibration observations & 320 LIBERO observations \\
Transformer blocks & 28 \\
Analyzed Linear families per block & 10 \\
Stream-aligned / text-context families & 8 / 2 \\
Temporal semantic streams & 9 \\
Denoising steps & 5, indexed 0--4 \\
Regions per observation & 10,360 \\
Weight quantization & Symmetric per-output-channel W4 \\
Activation quantization & Symmetric static per-tensor A8 \\
Action chunk & $16\times7$ \\
Rademacher projections & 16 \\
Denoising seed & 195 \\
Projection seed & 20260809 \\
\bottomrule
\end{tabular}
\end{table}

\subsection{Efficient Estimation of the Action-Impact Stream Map}
\label{app:map_estimation}

\textbf{Estimation target.}
For a fixed observation \(x\) and region \((\ell,\tau,s)\), let
\[
\mathbf{v}
=
\operatorname{vec}\!\left(\mathbf{E}_{\ell,\tau,s}(x)\right),
\qquad
\mathbf{J}
=
\mathbf{J}_{\ell,\tau,s}(x),
\]
where \(\mathbf{E}_{\ell,\tau,s}\) is the local joint
weight--activation quantization error defined in Section~\ref{sec:action_sensitivity_field}, and
\(\mathbf{J}_{\ell,\tau,s}\) is the Jacobian of the standardized
final action chunk with respect to
\(\operatorname{vec}(\mathbf{Y}_{\ell,\tau,s})\).
The first-order action response is \(\mathbf{Jv}\).
We estimate its mean squared magnitude,
\begin{equation}
e_{x,\ell,\tau,s}
=
\frac{1}{m}\|\mathbf{Jv}\|_2^2,
\label{eq:appendix_impact_target}
\end{equation}
where \(m=T d_a\), \(T\) is the action-chunk horizon, and
\(d_a\) is the action dimension.

\textbf{Random-projection estimator.}
Let \(\mathbf{z}_1,\ldots,\mathbf{z}_P\in\{-1,+1\}^{m}\)
be independent random vectors with independent Rademacher entries.
We estimate Eq.~\ref{eq:appendix_impact_target} by
\begin{equation}
\widehat{e}_{x,\ell,\tau,s}
=
\frac{1}{Pm}
\sum_{p=1}^{P}
\left\langle
\mathbf{J}^{\top}\mathbf{z}_p,\mathbf{v}
\right\rangle^2.
\label{eq:appendix_impact_estimator}
\end{equation}
Each term can be computed using a vector--Jacobian product (VJP)
because
\begin{equation}
\left\langle
\mathbf{J}^{\top}\mathbf{z}_p,\mathbf{v}
\right\rangle
=
\mathbf{z}_p^{\top}\mathbf{Jv}.
\label{eq:appendix_vjp_identity}
\end{equation}
Since
\(\mathbb{E}[\mathbf{z}_p\mathbf{z}_p^{\top}]=\mathbf{I}\),
\begin{equation}
\begin{aligned}
\mathbb{E}_{\mathbf{z}}
\left[\widehat{e}_{x,\ell,\tau,s}\right]
&=
\frac{1}{m}
\mathbf{v}^{\top}\mathbf{J}^{\top}
\mathbb{E}\!\left[\mathbf{z}_p\mathbf{z}_p^{\top}\right]
\mathbf{Jv}
\\
&=
\frac{1}{m}\|\mathbf{Jv}\|_2^2.
\end{aligned}
\label{eq:appendix_estimator_unbiasedness}
\end{equation}
Thus, the estimator is unbiased for the mean-squared first-order
action response.

We aggregate these estimates over the calibration set
\(\mathcal{D}\), with \(|\mathcal{D}|=N_{\mathrm{cal}}\), to obtain
\begin{equation}
\widehat{S}_{\ell,\tau,s}
=
\left[
\frac{1}{N_{\mathrm{cal}}}
\sum_{x\in\mathcal{D}}
\widehat{e}_{x,\ell,\tau,s}
\right]^{1/2}.
\label{eq:appendix_estimated_impact_score}
\end{equation}
For a fixed calibration set, unbiasedness applies to
\(\widehat{S}_{\ell,\tau,s}^{\,2}\), not generally to
\(\widehat{S}_{\ell,\tau,s}\).

\textbf{VJP reuse and computational cost.}
For each observation and projection, one reverse pass provides
VJPs at all captured Linear outputs along the full-precision
trajectory. At each output, we extract the VJP rows corresponding
to stream \(s\), vectorize them, and contract the result with
\(\operatorname{vec}(\mathbf{E}_{\ell,\tau,s})\) using
Eq.~\ref{eq:appendix_vjp_identity}.
For the text-context regions, the full output VJP is used.
Thus, the same reverse pass is reused across all regions.

Directly evaluating exact action displacements over \(R\) regions
requires \(\mathcal{O}(N_{\mathrm{cal}}R)\) downstream forward
evaluations. In contrast, the proposed estimator uses \(P=16\)
projections and requires \(\mathcal{O}(N_{\mathrm{cal}}P)\)
reverse passes.
Error construction and contractions still
scale with the number of regions, but require neither separate
downstream replays nor an explicit Jacobian.

\textbf{Local linearization error.}
The estimator above targets the first-order response.
To characterize its difference from the exact displacement, write
\[
\mathbf{y}
=
\operatorname{vec}\!\left(\mathbf{Y}_{\ell,\tau,s}(x)\right).
\]
For this fixed observation and region, let \(f(\mathbf{u})\) denote
the standardized final action chunk obtained by replacing the
selected stream's output with \(\operatorname{unvec}(\mathbf{u})\),
holding all unselected streams at their FP values at the injection
site, and running the remaining computation in full precision.
Conditioning and sampling randomness are held fixed.
Then \(\mathbf{J}=Df(\mathbf{y})\).

If the Jacobian \(Df\) is \(L_J\)-Lipschitz in the induced
Euclidean operator norm along the segment from \(\mathbf{y}\)
to \(\mathbf{y}+\mathbf{v}\), then
\begin{equation}
\left\|
f(\mathbf{y}+\mathbf{v})
-
f(\mathbf{y})
-
\mathbf{Jv}
\right\|_2
\leq
\frac{L_J}{2}\|\mathbf{v}\|_2^2.
\label{eq:appendix_linearization_bound}
\end{equation}
Under this smoothness assumption, the linearization error is
bounded by a term quadratic in the local quantization error
magnitude.

\subsection{Structure of the Action-Impact Stream Map}
\label{app:field_structure}

We analyze the Cosmos-Policy action-impact stream map constructed
from 320 LIBERO observations under the W4A8 reference configuration.
For a group of regions \(\mathcal{G}\), we aggregate the estimated
squared first-order action responses as
\begin{equation}
\widehat{S}_{\mathcal{G}}
=
\left[
\frac{1}{N_{\mathrm{cal}}|\mathcal{G}|}
\sum_{x\in\mathcal{D}}
\sum_{i\in\mathcal{G}}
\widehat{e}_{x,i}
\right]^{1/2},
\label{eq:marginal_field_aggregation}
\end{equation}
where \(i\) indexes a region \((\ell,\tau,s)\).
This quantity summarizes the action-impact scores of individual
regions; it does not estimate the combined action displacement
from simultaneously perturbing all regions in \(\mathcal{G}\).

\begin{figure}[t]
    \centering
    \includegraphics[width=\linewidth]
    {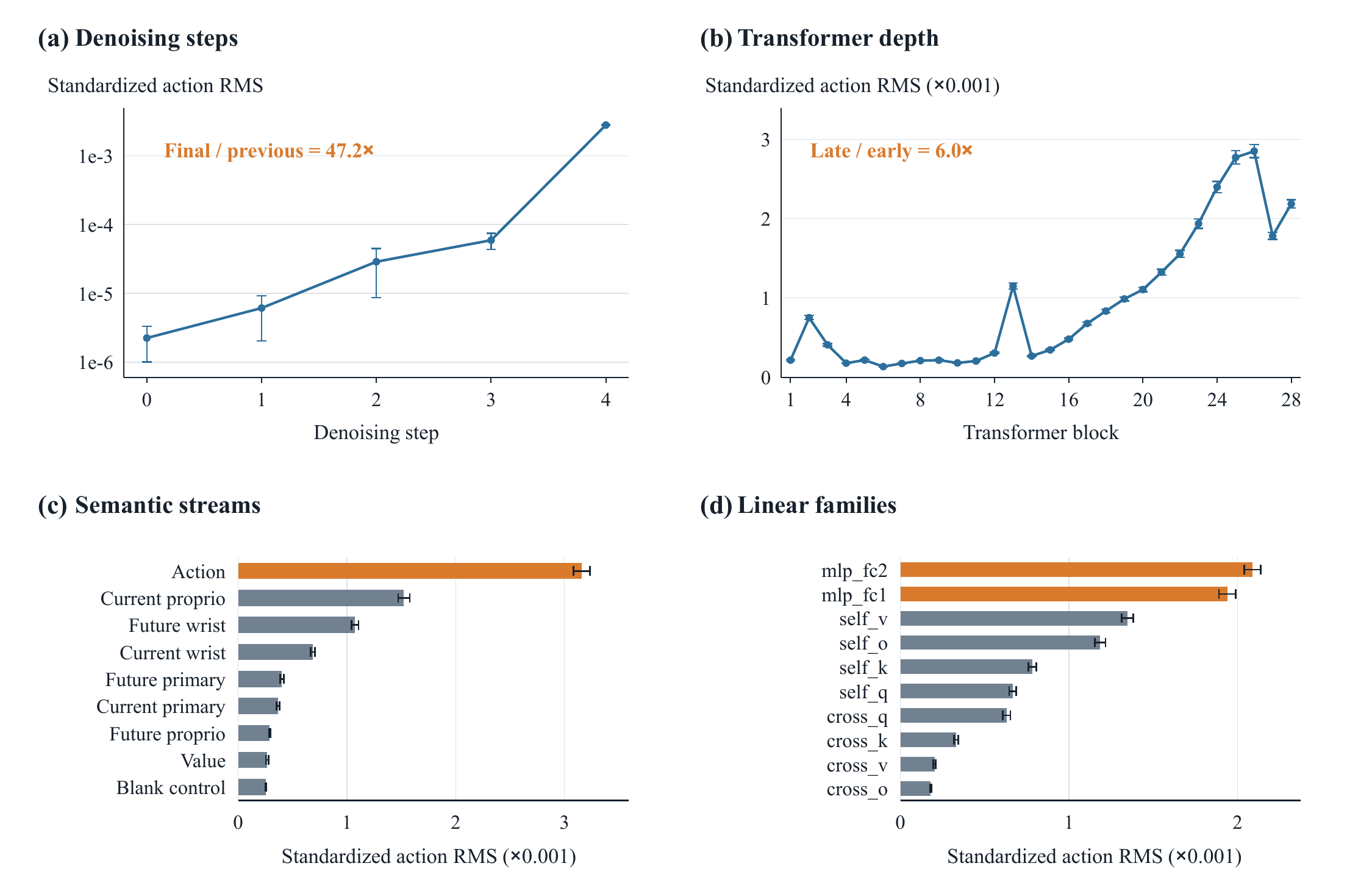}
    \caption{\textbf{Marginal action impact across four dimensions.}
(a) Denoising steps.
(b) Transformer blocks.
(c) Temporal semantic streams.
(d) Linear families.
Values are aggregated action-impact scores computed over
320 observations and the regions in each group using
Eq.~\ref{eq:marginal_field_aggregation}.
Panel (c) includes only the eight stream-aligned Linear families.
Error bars show 95\% observation-bootstrap intervals from
10,000 resamples.
The final/previous ratio compares Steps~4 and~3;
the late/early ratio compares Blocks~19--28 with Blocks~1--9.}
    \label{fig:map_marginals}
    \vspace{-8pt}
\end{figure}

\textbf{Marginal action impact.}
Figure~\ref{fig:map_marginals} summarizes the aggregated
action-impact scores across denoising steps, Transformer blocks,
semantic streams, and Linear families.
The final step has an aggregated score \(47.2\times\) that of
Step~3, while Blocks~19--28 have an aggregated score
\(6.0\times\) that of Blocks~1--9.
The action stream has the largest aggregated score among the
temporal semantic streams, and the two MLP projections have the
largest Linear-family scores.
These results show substantial differences in the estimated
first-order action impact of local quantization errors across
regions.

\begin{figure}[t]
    \centering
    \includegraphics[
        width=\linewidth,
        trim=45bp 305bp 70bp 35bp,
        clip
    ]{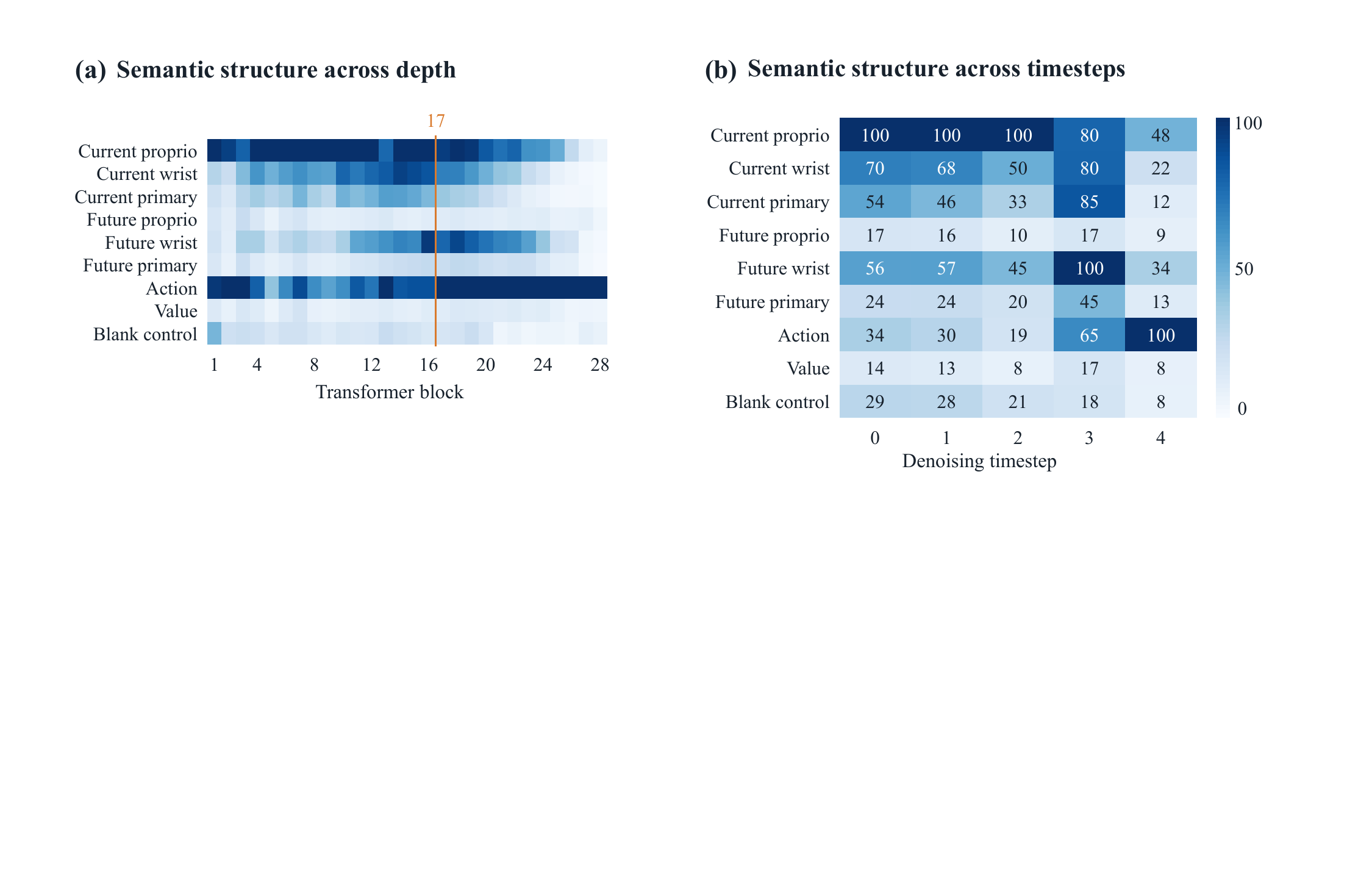}
    \caption{\textbf{Relative action impact of semantic streams.}
(a) Stream scores within each Transformer block, aggregated
over denoising steps and the eight stream-aligned Linear families.
(b) Stream scores within each denoising step, aggregated over
blocks and the eight stream-aligned Linear families.
Both panels aggregate over 320 observations using
Eq.~\ref{eq:marginal_field_aggregation}.
Each column is divided by its maximum and multiplied by 100,
so colors indicate relative action-impact scores within a column.
The marker at Block~17 identifies the first block from which
the action stream consistently has the largest block-wise
aggregated score.}
    \label{fig:map_semantic_structure}
    \vspace{-8pt}
\end{figure}

\textbf{Temporal and depth-dependent semantic structure.}
Figure~\ref{fig:map_semantic_structure} shows that the stream
with the largest aggregated action-impact score changes with
depth and denoising step.
After aggregation over denoising steps and the eight
stream-aligned Linear families, this stream varies across
earlier blocks but is consistently the action stream from
Block~17 onward.
After aggregation over blocks and the same eight Linear
families, current proprioception has the largest score at
Steps~0--2, future wrist image at Step~3, and action at Step~4.

\begin{figure}[!t]
    \centering
    \includegraphics[width=0.94\linewidth]
    {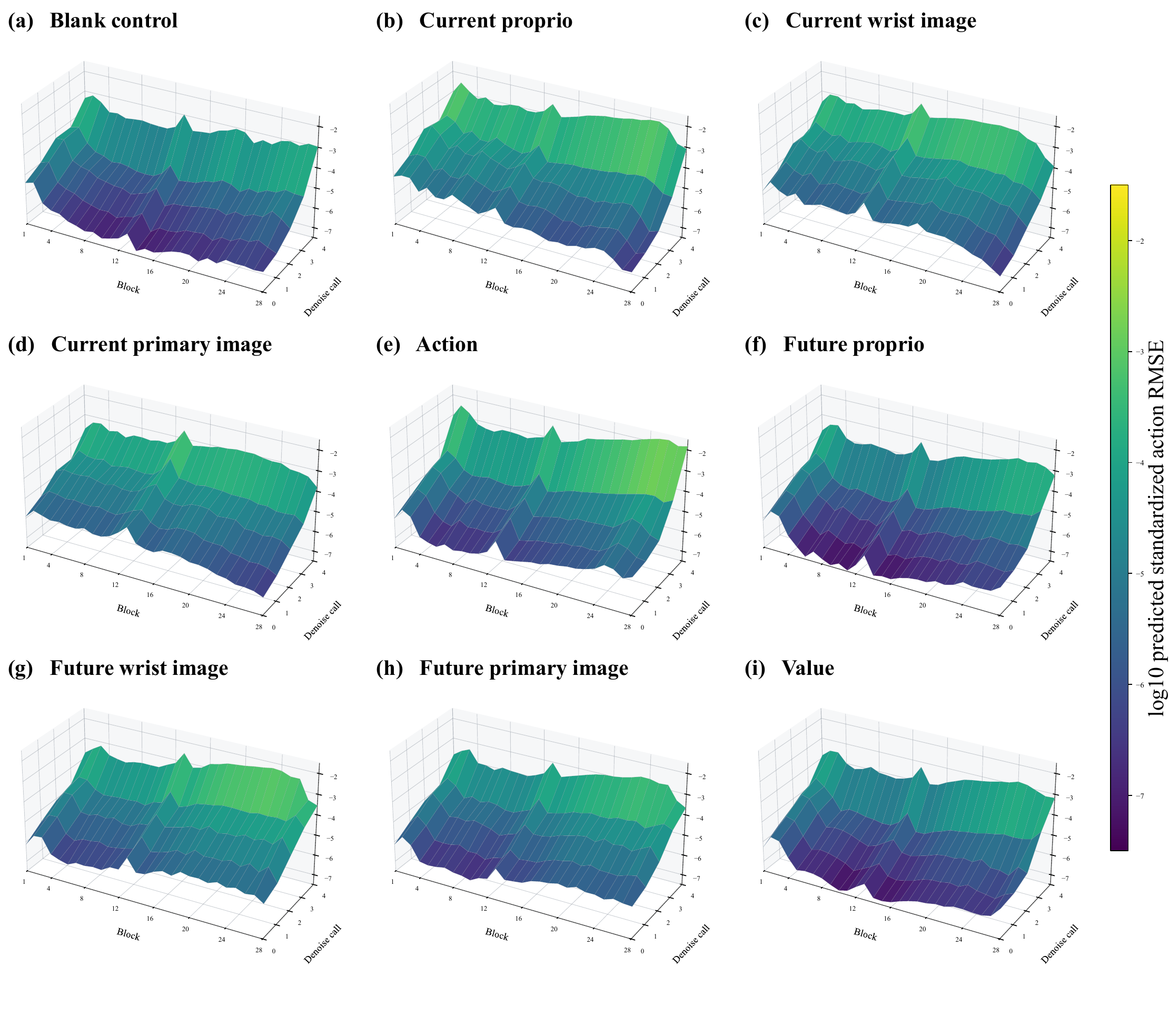}
    \caption{\textbf{Action-impact surfaces for nine temporal
semantic streams.}
Each panel shows the \(\log_{10}\) of the aggregated
action-impact score across 28 Transformer blocks and five
denoising steps.
At each block--step pair, scores are aggregated over the eight
stream-aligned Linear families and 320 observations using
Eq.~\ref{eq:marginal_field_aggregation}.
Surface height and color encode the same quantity.
All panels share the same vertical and color scales.}
    \label{fig:map_surfaces}
    \vspace{-8pt}
\end{figure}

Figure~\ref{fig:map_surfaces} complements these summaries by
showing the joint depth--step distribution of action-impact
scores for each of the nine temporal streams on shared scales.
Together, these figures reveal variation in the estimated
action impact of quantization errors across Linear layers,
denoising steps, and semantic streams, motivating calibration
that prioritizes regions with greater impact on final actions.

\subsection{Stability of the Estimated Map}
\label{app:field_stability}

\begin{table}[t]
    \centering
    \setlength{\belowcaptionskip}{5pt}
    \caption{
    Agreement with the 320-observation reference map
    as calibration size increases.
    Entries report medians and 2.5th--97.5th percentile
    intervals over 200 subsets at each size.
    Observations are sampled without replacement
    within each subset.
    }
    \label{tab:field_convergence}
    \small
    \vspace{-5pt}
    \begin{tabular}{@{}ccc@{}}
        \toprule
        Observations & Spearman correlation & Top-10\% overlap \\
        \midrule
        5   & 0.898 [0.859, 0.958] & 0.957 [0.914, 0.965] \\
        10  & 0.927 [0.864, 0.969] & 0.966 [0.929, 0.973] \\
        20  & 0.947 [0.874, 0.981] & 0.973 [0.954, 0.979] \\
        40  & 0.960 [0.911, 0.987] & 0.980 [0.972, 0.984] \\
        80  & 0.977 [0.938, 0.993] & 0.985 [0.981, 0.988] \\
        160 & 0.993 [0.970, 0.999] & 0.989 [0.986, 0.992] \\
        \bottomrule
    \end{tabular}
\end{table}

We assess the stability of regional action-impact rankings
under changes in calibration-sample selection, with the
reference quantization configuration and map-estimation
settings held fixed.

\textbf{Split-half agreement.}
We randomly divide the 320 observations into two disjoint
subsets of 160 and estimate a map from each using
Eq.~\ref{eq:appendix_estimated_impact_score}.
Across 200 partitions, the median Spearman correlation
over all 10,360 regions is \(0.967\)
(2.5th--97.5th percentiles: \(0.953\)--\(0.974\)).
The median top-10\% overlap is \(0.979\)
(2.5th--97.5th percentiles: \(0.974\)--\(0.984\)),
where overlap is the intersection size of the two
top-ranked sets, each containing 1,036 regions,
divided by 1,036.
Median correlations computed separately within each
denoising step range from \(0.890\) to \(0.999\).
Thus, agreement remains high both across the full map
and within individual denoising steps.

\textbf{Calibration-size convergence.}
For each sample size in Table~\ref{tab:field_convergence},
we draw 200 subsets, sampling observations without
replacement within each subset, and compare their
estimated maps with the 320-observation reference.
We use the same Spearman correlation and top-10\%
overlap metrics as in the split-half analysis.
Median agreement increases with sample size, and even
small subsets recover most of the highest-impact regions
in the reference map.
Because the subsets are contained in the reference pool,
this analysis measures convergence to the full-pool
estimate, complementing the disjoint split-half comparison.

Together, these results show stable regional action-impact rankings
across calibration subsets, supporting the use of a fixed map to
guide subsequent calibration.

\section{Calibration Weights and Activation Constraints}
\label{app:field_aggregation}

\subsection{From Action-Impact Scores to Calibration Weights}

Let $\widehat S_{\ell,\tau,s}$ denote the estimated action-impact
score. We convert its square into a positive calibration weight
before normalization:
\begin{equation}
u_{\ell,\tau,s}
=
\left(\widehat S_{\ell,\tau,s}^{\,2}+\eta\right)^\rho,
\qquad
\eta>0,\quad 0<\rho\le1.
\label{eq:sensitivity_temperature}
\end{equation}
The constant $\eta$ ensures positive weights, while $\rho<1$
reduces the dominance of large scores.

Since $\mathbf D_\ell$ is shared across denoising steps and
streams, its weights are normalized jointly over both:
\begin{equation}
\omega^D_{\ell,\tau,s}
=
\frac{u_{\ell,\tau,s}}
{\sum_{\tau',j}\pi_{\tau'}u_{\ell,\tau',j}},
\qquad
\sum_{\tau,s}\pi_\tau\omega^D_{\ell,\tau,s}=1.
\label{eq:static_sensitivity_weight}
\end{equation}
Here, $\pi_\tau$ is the normalized calibration frequency
of step $\tau$, with $\sum_\tau\pi_\tau=1$.

Activation calibration operates separately at each layer
and step, so its weights are normalized over streams:
\begin{equation}
\omega^\gamma_{\ell,\tau,s}
=
\frac{u_{\ell,\tau,s}}
{\sum_j u_{\ell,\tau,j}},
\qquad
\sum_s\omega^\gamma_{\ell,\tau,s}=1.
\label{eq:dynamic_sensitivity_weight}
\end{equation}
The estimated map and both sets of weights remain fixed
throughout calibration.

\subsection{Effective Activation Ranges and Constraints}
\label{app:activation_constraints}

For a fixed layer and denoising step, inverse-gain compensation
gives stream $s$ an effective clipping threshold $c/\gamma_s$
and quantization step size $c/(q_{\max}\gamma_s)$.
Here, $q_{\max}$ corresponds to the activation bit-width.
Increasing $\gamma_s$ therefore reduces rounding intervals
while narrowing the unclipped range.

All streams share the same gain bounds, which restrict their
effective thresholds to
\begin{equation}
\frac{c}{\gamma_s}
\in
\left[
\frac{c}{\gamma_{\max}},
\frac{c}{\gamma_{\min}}
\right].
\end{equation}
The normalization $\sum_s n_s\log\gamma_s=0$ implies
\begin{equation}
\prod_s
\left(\frac{c}{\gamma_s}\right)^{
n_s/\sum_j n_j}
=
c.
\end{equation}
Thus, $c$ is the token-weighted geometric mean of the effective
thresholds. Together with the bounds on $c$ and $\gamma_s$,
this restricts the combinations of quantization ranges
available across streams.
When no feasible configuration attains every stream's
individually preferred threshold, the map-weighted objective
guides the trade-off among them.

The reconstruction target
$\widetilde{\mathbf X}_{\ell,\tau,s}\widehat{\mathbf W}_\ell$
uses unquantized activations and fixed quantized weights,
isolating the additional error from activation quantization.
After calibration, all streams use the shared base step size
$c_{\ell,\tau}/q_{\max}$ before inverse-gain compensation,
which produces their stream-specific effective step sizes.

\begingroup
\subsection{Sequential Optimization and Parameter Selection}
\label{app:optimization}
All map-derived weights are computed once and remain fixed. The same calibration observations generate inputs across denoising steps and streams; separate datasets are not collected for individual regions. We cache layer inputs and compute reference outputs during optimization. Both stages use Adam in log space to preserve positivity and a straight-through estimator through rounding.

\textbf{Routing.} For each layer, update only $\mathbf D_\ell$ in Eq.~\ref{eq:static_calibration_objective}, holding the original model weights fixed. Recompute $\mathcal Q(\mathbf D_\ell\mathbf W_\ell)$ as the scales change. Accumulate gradients across denoising steps and streams before each of 20 Adam updates, with learning rate 0.02. Select the best candidate and freeze $\mathbf D_\ell^*$ and $\widehat{\mathbf W}_\ell=\mathcal Q(\mathbf D_\ell^*\mathbf W_\ell)$.

\textbf{Modulation.} At each layer and denoising step, jointly optimize stream gains and a shared threshold using 30 Adam updates, with learning rates 0.02 for gain parameters and 0.01 for the log threshold. The target $\widetilde{\mathbf X}\widehat{\mathbf W}_\ell$ isolates activation-quantization error. With $p_s=n_s/\sum_j n_j$, center raw log gains by their $p_s$-weighted mean and apply a common shrink factor before exponentiation. This preserves $\sum_s n_s\log\gamma_s=0$ and bounds $\gamma_s\in[0.25,4]$. After each update, clamp $\log c$ to $[\log(0.01c_0),\log(4c_0)]$, where $c_0$ is the initial threshold.

\textbf{Hard-quantization selection.} In both stages, evaluate the hard-quantization loss over all cached rows every five updates and at the final update. Retain the lowest-loss candidate, including initialization, before advancing to the next stage or inference. No online parameter optimization is performed.

\par\endgroup

\section{Additional Experimental Details}
\label{app:experimental_details}

\subsection{Quantization Settings}

\begin{figure*}[t]
\centering
\includegraphics[
    width=\textwidth,
    trim={1.0cm 2.7cm 0.9cm 3.1cm},
    clip
]{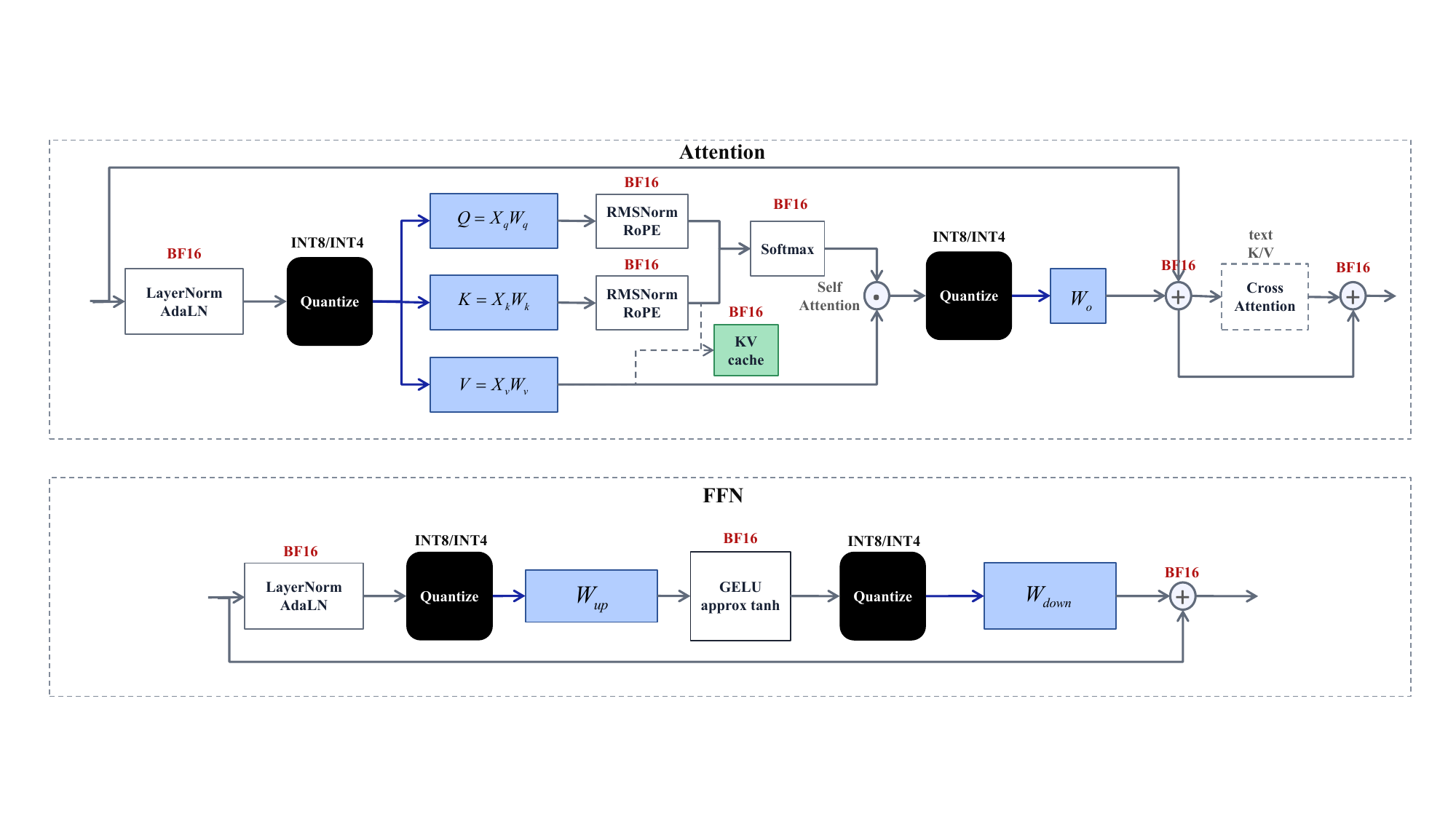}
\caption{\textbf{Transformer linear-layer quantization scope.}
The same quantization scope is used for Cosmos-Policy,
FastWAMJoint, and Cosmos-Edge.}
\label{fig:quantization_scope}
\vspace{-8pt}
\end{figure*}

\textbf{Quantization scope.}
We apply the same quantization scope to Cosmos-Policy,
FastWAMJoint, and Cosmos-Edge.
We quantize the query, key, value, and output projections
of both self-attention and cross-attention, together
with the FFN up- and down-projection layers.
Their weights use 4-bit precision, while their input
activations use 8-bit or 4-bit precision under W4A8
or W4A4, respectively.
Normalization, RoPE, Softmax, GELU, the KV cache, and residual additions remain in BF16
(Figure~\ref{fig:quantization_scope}).

\textbf{Quantization format.}
For \methodname{}, weights are quantized symmetrically
per output channel. Activations are quantized symmetrically
using an offline-calibrated base scale shared across streams
at each layer and denoising step. Stream gains and inverse-gain
compensation produce stream-specific effective scales.


\subsection{Calibration and Evaluation}
\label{app:robolab_calibration}
\textbf{Calibration data.}
We sample 320 observations from the LIBERO training set
{and 192 from the RoboCasa training set}
for Cosmos-Policy, 100 from the LIBERO training set
for FastWAMJoint, and 512 from the DROID~\citep{khazatsky2024droid} training set
for Cosmos-Edge.
Within each model {and benchmark}, all methods use the same calibration set.

{\textbf{RoboLab results.}
Our BF16 Cosmos-Edge achieves 21.83\% success on RoboLab.
NVIDIA reports \href{https://developer.nvidia.com/blog/post-train-nvidia-cosmos-3-edge-for-on-device-robot-control/}{22.9\%}
for its DROID-trained Edge policy under its own RoboLab protocol.
W4A8 and W4A4 \methodname{} fall 7.16 and 4.50 percentage points
below BF16, respectively. RoboLab places greater emphasis on multi-step
tasks than DROID~\citep{yang2026robolab}, so the DROID-calibrated ranges
may cover activations encountered on RoboLab less well. We have not
isolated this effect or established why W4A4 exceeds W4A8 by 2.66 points.}

\textbf{Evaluation data.}
Cosmos-Policy is evaluated {on LIBERO} over 3,000 rollouts,
covering 40 tasks with 25 trials per task
under each of three random seeds.
{On RoboCasa, Cosmos-Policy is evaluated over 3,600 rollouts,
covering 24 tasks with 50 trials per task under each of three random seeds.}
FastWAMJoint and Cosmos-Edge are evaluated over
2,000 and 1,200 rollouts, respectively{, in total across the three evaluation seeds}.
The calibration and evaluation data are non-overlapping.

\textbf{Evaluation protocol.}
All methods use matched initial states and random seeds.
We report the mean closed-loop success rate across tasks.
For ablations, standardized action RMSE is computed
relative to full-precision predictions using fixed
training-set statistics, with matched observations
and sampling randomness.

\textbf{Efficiency measurements.}
All three models follow the same general benchmarking procedure
on an NVIDIA RTX 4090 GPU.
We measure DiT replay latency using FP-captured inputs,
CUDA Graphs, and CUDA events, with warm-up runs before timing.
For Cosmos-Policy, we use five warm-up sequences and 20 timed
sequences and report the median. Speedup is computed relative to each model's BF16 reference.

\textbf{Real-world evaluation.}
We evaluate FastWAMJoint fine-tuned on the real-world tasks,
using the same checkpoint for BF16 and W4A8 SteerQuant.
Each method is evaluated over 12 trials per task under matched
initial conditions.
Pack Objects receives a score of $k/3$, where $k$ is the number
of successfully handled objects; the other tasks use binary
success scores. We report each task's mean score over 12 trials as a percentage.
\begingroup
\section{Additional Task Evaluations and Calibration Analysis}
\label{app:revision_experiments}
\par\endgroup

\begingroup
\subsection{RoboCasa Evaluation}
\label{app:robocasa}
Cosmos-Policy is evaluated on 24 RoboCasa tasks with 50 rollouts per task under each of three evaluation seeds, giving 3,600 rollouts per method. Success rates are reported as the mean and sample SD across seeds. Table~\ref{tab:robocasa_full} jointly reports success, DiT latency, peak memory and packed storage. PTQ4DiT and Q-DiT lack low-bit inference kernels for the evaluated settings, so their efficiency entries are unavailable. Latency is the five-step DiT replay, not the entire policy request. The main benchmark uses an RTX 4090, batch size one, FP-captured inputs, CUDA Graphs and CUDA-event timing, with five warm-up sequences and 20 timed sequences, reporting the median. The separately controlled QuaRot comparison in Appendix~\ref{app:quarot_control} uses a different repeated-process protocol; its absolute latencies must not be pooled with this table.

\par\endgroup
\begin{table}[htbp]
\centering

\caption{\textbf{Cosmos-Policy on RoboCasa.} SR is mean $\pm$ sample SD across three seeds. Speedup is relative to BF16. Peak memory and packed storage use GB and GiB, respectively.}\label{tab:robocasa_full}
\footnotesize
\setlength{\tabcolsep}{4pt}\renewcommand{\arraystretch}{1.08}
\begin{tabular}{@{}lcccccc@{}}
\toprule
Precision & Method & SR (\%) & Latency (ms) & Speedup & Peak (GB) & Storage (GiB) \\
\midrule
FP & Original & $63.33\pm0.72$ & 366.100 & 1.000 & 4.742 & 3.644 \\
W4A8 & SmoothQuant & $54.30\pm0.87$ & 242.570 & 1.509 & 2.109 & 1.189 \\
 & PTQ4DiT & $58.61\pm1.05$ & N/A & N/A & N/A & N/A \\
 & Q-DiT & $57.92\pm0.84$ & N/A & N/A & N/A & N/A \\
 & SteerQuant & $63.20\pm2.55$ & 239.886 & 1.526 & 2.233 & 1.189 \\
W4A4 & QuaRot & $54.59\pm2.60$ & 198.522 & 1.844 & 2.106 & 1.189 \\
 & Atom & $53.05\pm0.87$ & 312.327 & 1.172 & 2.184 & 1.258 \\
 & SVDQuant & $50.83\pm1.50$ & 231.204 & 1.583 & 2.283 & 1.320 \\
 & SteerQuant & $60.14\pm1.97$ & 188.817 & 1.939 & 2.236 & 1.191 \\
\bottomrule
\end{tabular}
\end{table}
\begingroup
\subsection{Video-only and Fixed-W4 QuantWAMs Baselines}
\label{app:additional_baselines}
\textbf{Video-only activation quantization.} All target Linear weights remain shared and packed in INT4. Video activations use A8 or A4; non-video activations, including proprioceptive and action streams, remain BF16. Low-bit video processing and BF16 non-video computation require conversion and separate execution paths in the evaluated implementation. These costs can outweigh the savings from quantization. At W4A4, retaining BF16 non-video activations is insufficient to preserve task success; errors in video processing can still affect actions through shared computation.

\textbf{QuantWAMs-FixedW4.} We adapt QuantWAMs~\citep{zhou2026quantwams} using the same 320-window calibration protocol, fixing all 280 target Linear layers to packed INT4. We retain rotation and smoothing, shared-mask calibration, joint video--action saliency, GPTQ compensation and rollout-based denoising-step protection. A8 is retained in 56/280 layers and at the fifth of five steps, with approximately 2\% BF16 exception activations. The exception branch also uses W4 weights. These protections yield a nominal activation precision of 5.65 bits when target layers and denoising steps are weighted equally, compared with four bits for SteerQuant's target Linear activations. This is a fixed-weight-precision comparison, not an identical-activation-precision comparison.

At fixed W4, QuantWAMs has less freedom to allocate weight precision. SteerQuant instead calibrates shared channel scales and stream-specific activation ranges, with fused execution. In this setting, SteerQuant improves SR from 95.60\% to 98.30\%, reduces DiT latency by 30.71\% and reduces peak memory by 40.16\%.

\par\endgroup

\begingroup
\subsection{Full Recalibration across Calibration Subsets}
\label{app:subset_recalibration}
Appendix~\ref{app:field_stability} evaluates map-ranking agreement, whereas this experiment repeats Map estimation, Routing and Modulation on three subsets of 40 observations and three subsets of 160 observations. All models use W4A8 and are evaluated on the same 900 observations with matched FP references for standardized action RMSE. Closed-loop evaluation uses 60 episodes over 20 tasks from LIBERO-Spatial and LIBERO-10, with three matched initial states per task and a fixed policy seed. The SD in Table~\ref{tab:subset_recalibration} is across calibration subsets, not evaluation seeds. The 320-observation row is the original calibrated reference. These results characterize sensitivity within the tested calibration pool rather than distribution shift.

\par\endgroup
\begin{table}[htbp]
\centering

\caption{\textbf{Sensitivity to calibration sample size.} Mean $\pm$ sample SD across three selected subsets at each smaller size; the full-pool reference has no subset SD.}\label{tab:subset_recalibration}
\small
\setlength{\tabcolsep}{4pt}\renewcommand{\arraystretch}{1.08}
\begin{tabular}{@{}lcc@{}}
\toprule
Observations & Standardized action RMSE & Closed-loop SR (\%) \\
\midrule
40 & $0.12493\pm0.00524$ & $94.44\pm0.96$ \\
160 & $0.12197\pm0.00103$ & $97.22\pm0.96$ \\
320 (reference) & 0.12314 & 98.33 \\
\bottomrule
\end{tabular}
\end{table}
\begingroup
\subsection{Offline Calibration Time and Memory}
\label{app:calibration_cost}
We measure Cosmos-Policy W4A8 calibration with 320 LIBERO observations on one NVIDIA A100-SXM4-80GB. Times include model loading, data collection, optimization and saving, but exclude evaluation. Peak allocated GPU memory includes the resident model. Routing and Modulation times are accumulated across all 280 layers. Complete calibration takes 298.03 minutes (4.967 h) and peaks at 35.56 GiB. Map estimation uses 16 projections per observation, reusing reverse passes across regions without dense Jacobians; it accounts for 73.2\% of calibration time. Routing and Modulation together only take 41.36 minutes. The map's computational cost is offline and does not imply online optimization. Table~\ref{tab:calibration_comparison} compares calibration times under matched experimental conditions; SteerQuant is faster than QuaRot, SVDQuant, PTQ4DiT and Q-DiT, but slower than SmoothQuant and Atom.

\par\endgroup
\begin{table}[htbp]
\centering

\caption{\textbf{Offline calibration stages.} Peak values are allocated GPU memory, including the resident model.}\label{tab:calibration_stages}
\small
\setlength{\tabcolsep}{4pt}\renewcommand{\arraystretch}{1.08}
\begin{tabular}{@{}lcc@{}}
\toprule
Stage & Time (min) & Peak GPU memory (GiB) \\
\midrule
Activation statistics & 6.21 & 5.06 \\
Map estimation & 218.05 & 35.56 \\
Activation caching & 22.71 & 5.06 \\
Routing & 17.23 & 4.64 \\
Modulation & 24.13 & 4.29 \\
Loading, saving and other overhead & 9.70 & -- \\
Complete calibration & 298.03 & 35.56 \\
\bottomrule
\end{tabular}
\end{table}
\begin{table}[htbp]
\centering

\caption{\textbf{Calibration time under matched conditions.}}\label{tab:calibration_comparison}
\small
\setlength{\tabcolsep}{4pt}\renewcommand{\arraystretch}{1.08}
\begin{tabular}{@{}lc@{}}
\toprule
Method & Calibration time (h) \\
\midrule
SmoothQuant & 0.617 \\
QuaRot & 5.317 \\
Atom & 3.750 \\
SVDQuant & 7.340 \\
PTQ4DiT & 15.140 \\
Q-DiT & 14.330 \\
SteerQuant & 4.967 \\
\bottomrule
\end{tabular}
\end{table}
\begingroup
\section{Controlled Execution and Hardware Analysis}
\label{app:execution_analysis}
\par\endgroup

\begingroup
\subsection{Ablation of Rudder's Execution Optimizations}
\label{app:matched_execution}

We evaluate Cosmos-Policy W4A8 on an RTX 4090 at batch
size one, using matched BF16-captured LIBERO inputs for
five denoising steps.
Configurations A/B/C share identical calibrated parameters,
packed weights and GEMM workload.
A materializes channel scaling; B fuses channel scaling
into activation quantization; C additionally enables
block-level normalization/AdaLN and gate/residual fusion.
Stream gains and output compensation stay fused in all
three configurations.

We time five-step DiT replay using CUDA Graphs and CUDA
events after warm-up, over 10 independent processes with
30 randomized, interleaved repetitions per process.
Table~\ref{tab:rudder_execution} reports the median of
process medians.
Timing excludes model loading, compilation, graph construction
and all policy stages outside the DiT.
Kernel counts come from separate profiling of one
representative input.

Full Rudder achieves a 31.97\% mean paired latency reduction
relative to A; block-level fusion provides a further 7.32\%
mean paired reduction relative to B.
These paired statistics are computed across processes.
C is faster than A and B in every process, removing
5,040 kernel executions while retaining the same
1,400 integer GEMM calls.
A/B produce bitwise-identical outputs; C changes intermediate
floating-point rounding.
Under this protocol, C achieves a $1.47\times$ speedup over A (205.87 vs.\ 140.05 ms).
The 133.455-ms value in Table~\ref{tab:efficiency} uses
the separate main benchmark with five warm-up and
20 timed sequences.
\par\endgroup

\begin{table}[htbp]
\centering

\caption{\textbf{Matched Rudder execution ablation.}
One five-step DiT replay with identical GEMM work
and calibrated parameters.}
\label{tab:rudder_execution}
\footnotesize
\setlength{\tabcolsep}{4pt}
\renewcommand{\arraystretch}{1.08}
\begin{tabular}{@{}lcccc@{}}
\toprule
Configuration & Scaling fusion & Block fusion
& Latency (ms) & Kernel executions \\
\midrule
A: materialized scaling
& $\times$ & $\times$ & 205.87 & 12,045 \\
B: scaling fusion
& $\checkmark$ & $\times$ & 151.13 & 9,245 \\
C: full Rudder
& $\checkmark$ & $\checkmark$ & 140.05 & 7,005 \\
\bottomrule
\end{tabular}
\end{table}

\begingroup
\subsection{Controlled Comparisons with Quantization Baselines}
\label{app:controlled_baselines}
\label{app:quarot_control}

\textbf{SmoothQuant.}
SmoothQuant and SteerQuant share the CUTLASS backend
in the W4A8 comparison on matched LIBERO inputs.
Ten independent processes with 30 randomized, interleaved
repetitions per process show lower SteerQuant latency
in every process.
The mean saving per five-step sequence is 1.838 ms
(95\% CI [1.794, 1.882] ms; 1.20\%).
Offline-calibrated thresholds eliminate dynamic per-token
threshold estimation in the evaluated SmoothQuant
implementation, with profiling showing reduced
activation-quantization cost.

\textbf{QuaRot.}
For Cosmos-Policy W4A4 on RoboCasa, we evaluate a QuaRot
DiT adaptation retaining applicable official optimizations
and a second adaptation adding Hadamard--quantization fusion
and GEMM-epilogue output dequantization.
Both adaptations and SteerQuant use identical captured inputs,
an RTX 4090, batch size one, CUDA Graphs and CUDA-event
timing.
Each implementation is measured in three independent
processes with 30 timed repetitions and cyclically
balanced process order.
Table~\ref{tab:quarot_control} reports the median of
the three process medians.
SteerQuant reduces latency by 40.36\% and 9.24\%
relative to the two adaptations, respectively.
Its execution includes additional normalization/AdaLN
and gate/residual fusion.
These measurements use a separate controlled protocol
from the main benchmark in Table~\ref{tab:robocasa_full}.
\par\endgroup

\begin{table}[htbp]
\centering

\caption{\textbf{Controlled QuaRot comparison on RoboCasa.}
Latency covers five DiT denoising steps.
Speedup is relative to the first row.}
\label{tab:quarot_control}
\small
\setlength{\tabcolsep}{4pt}
\renewcommand{\arraystretch}{1.08}
\begin{tabular}{@{}lcc@{}}
\toprule
Method & Latency (ms) & Speedup \\
\midrule
QuaRot: applicable official optimizations
& 287.437 & $1.000\times$ \\
QuaRot: additional fusion
& 188.877 & $1.522\times$ \\
SteerQuant
& 171.419 & $1.677\times$ \\
\bottomrule
\end{tabular}
\end{table}

\begingroup
\subsection{Bandwidth, Arithmetic Throughput and Roofline Estimates}
\label{app:roofline}

We combine Nsight Systems profiling with an empirical roofline
model on matched five-step LIBERO inputs.
Across the 280 quantized linear layers, the GEMM workload
remains 29.60 trillion operations at BF16, W4A8 and W4A4.
The gains come from reduced data movement and higher low-bit
arithmetic throughput, with the operation count unchanged.
Table~\ref{tab:roofline} separates modeled bandwidth- and
compute-limited times from measured GEMM times.

The model assumes one read of each matrix operand and one
BF16 output write, using measured streaming bandwidth and
large-matrix throughput as references.
It gives $1.87\times/2.15\times$ memory-side potential and
$3.43\times/6.57\times$ compute-side potential for W4A8/W4A4.
These describe separate limiting cases: using aggregate
workload totals, the roofline estimate is set by the larger
of data-transfer time and compute time,
$t \approx \max(D/B, F/P)$,
where $D$ is the operand payload, $B$ is the bandwidth,
$F$ is the operation count, and $P$ is the arithmetic throughput.
The two potentials therefore cannot be added as independent
speedup contributions.
The measured GEMM-stage speedups are $2.94\times$ and
$5.15\times$, respectively.

\par\endgroup
\begin{table}[htbp]
\centering

\caption{\textbf{GEMM workload analysis.} Modeled transfer/compute times and measured GEMM times are distinct quantities. Speedups refer to the GEMM stage only.}\label{tab:roofline}
\small
\setlength{\tabcolsep}{4pt}\renewcommand{\arraystretch}{1.08}
\begin{tabular}{@{}lccc@{}}
\toprule
Metric & BF16 & W4A8 & W4A4 \\
\midrule
Operand payload (GB, model) & 40.75 & 21.83 & 18.97 \\
Bandwidth-limited time (ms, model) & 44.15 & 23.65 & 20.55 \\
Compute-limited time (ms, model) & 183.12 & 53.39 & 27.89 \\
GEMM time (ms, Nsight Systems) & 199.29 & 67.80 & 38.69 \\
Roofline-estimated GEMM speedup & 1.00 & 3.43 & 6.57 \\
Measured GEMM speedup & 1.00 & 2.94 & 5.15 \\
\bottomrule
\end{tabular}
\end{table}
\begingroup
\subsection{Additional Rudder Storage and Arithmetic}
\label{app:rudder_overhead}
The allocated reshaping state, including retained copies, is 5.80 MiB, or 0.69\% of packed weights. The additional per-linear temporary allocation peak is zero for all four observed shapes. Channel-scaling multiplications add 0.0193\% of the GEMM operation count; this excludes per-row work and is not a count of all additional arithmetic. These allocation measurements describe storage requirements, not measured memory traffic.

Measured GEMM speedups are $2.94\times$ and $5.15\times$, about 86\% and 78\% of the corresponding roofline estimates. The overall gap includes utilization and execution effects and cannot be attributed entirely to Rudder.

\par\endgroup
\begin{table}[htbp]
\centering

\caption{\textbf{Additional Rudder costs.} Limits of each measurement are stated explicitly.}\label{tab:rudder_overhead}
\footnotesize
\setlength{\tabcolsep}{4pt}\renewcommand{\arraystretch}{1.08}
\begin{tabular}{@{}lc@{}}
\toprule
Metric & Measured result \\
\midrule
Allocated reshaping state (including retained copies) & 5.80 MiB; 0.69\% of packed weights \\
Additional per-linear temporary peak & 0 across four observed shapes \\
Channel-scaling multiplications / GEMM operations & 0.0193\%, excluding per-row work \\
\bottomrule
\end{tabular}
\end{table}

\end{document}